\documentclass[runningheads]{llncs}
\usepackage[T1]{fontenc}
\usepackage{graphicx}
\usepackage{amsmath,amssymb} 
\usepackage{color}
\usepackage[width=122mm,left=12mm,paperwidth=146mm,height=193mm,top=12mm,paperheight=217mm]{geometry}
\newcommand\blfootnote[1]{%
  \begingroup
  \renewcommand\thefootnote{}\footnote{#1}%
  \addtocounter{footnote}{-1}%
  \endgroup
}

\begin{document}
\pagestyle{headings}
\setcounter{secnumdepth}{3}
\mainmatter
\def\ECCV18SubNumber{1442}  

\title{Contextual-based Image Inpainting: Infer, Match, and Translate} 

\titlerunning{Contextual-based Image Inpainting: Infer, Match, and Translate}

\authorrunning{Y. Song et al.}

\author{Yuhang Song*\inst{1} \and
	    Chao Yang*\inst{1} \and
	    Zhe Lin\inst{2} \and
	    Xiaofeng Liu \inst{3} \and
	    Qin Huang \inst{1} \and
	    Hao Li \inst{1,4,5} \and
	    C.-C. Jay Kuo \inst{1}
}
\institute{USC, \email{\{yuhangso,chaoy,qinhuang\}@usc.edu,cckuo@sipi.usc.edu}, \and
		   Adobe Research, \email{zlin@adobe.com}, \and
		   CMU, \email{liuxiaofeng@cmu.edu}, \and
		   Pinscreen,\email{hao@hao-li.com}, \and
		   USC Institute for Creative Technologies
}

\maketitle
\blfootnote{* indicates equal contribution}
\begin{abstract}
We study the task of image inpainting, which is to fill in the missing region of an incomplete image with plausible contents. To this end, we propose a learning-based approach to generate visually coherent completion given a high-resolution image with missing components. In order to overcome the difficulty to directly learn the distribution of high-dimensional image data, we divide the task into inference and translation as two separate steps and model each step with a deep neural network. We also use simple heuristics to guide the propagation of local textures from the boundary to the hole. We show that, by using such techniques, inpainting reduces to the problem of learning two image-feature translation functions in much smaller space and hence easier to train. We evaluate our method on several public datasets and show that we generate results of better visual quality than previous state-of-the-art methods. 
\end{abstract}

\section{Introduction}

\begin{figure}
\centering
\small
\setlength{\tabcolsep}{1pt}
\begin{tabular}{cccccc}
  \includegraphics[width=.16\textwidth]{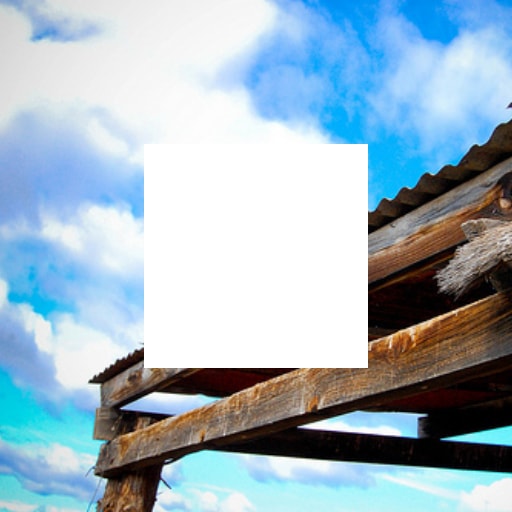}&
  \includegraphics[width=.16\textwidth]{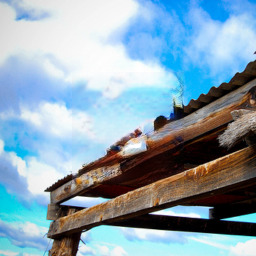}&
  \includegraphics[width=.16\textwidth]{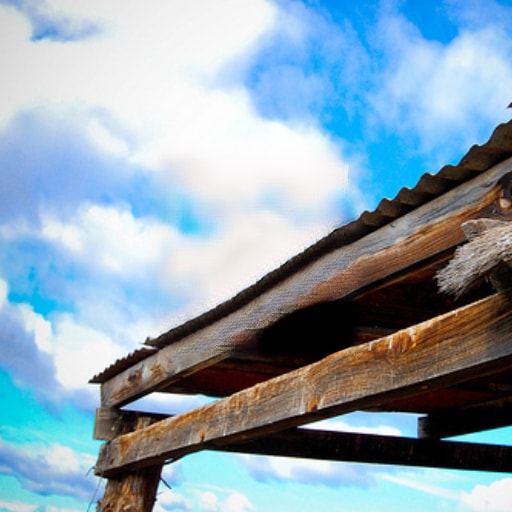}&
   \includegraphics[width=.16\textwidth]{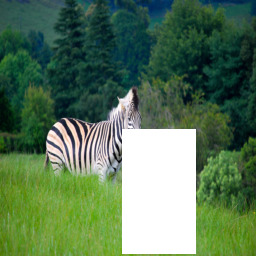}&
  \includegraphics[width=.16\textwidth]{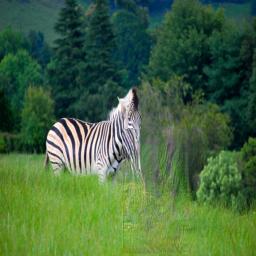}&
 \includegraphics[width=.16\textwidth]{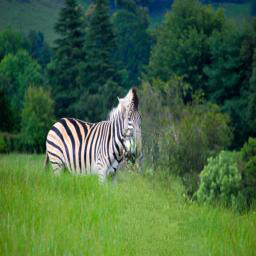}\\
(a) & (b) & (c) & (d) & (e) & (f) \\
\end{tabular}
\caption{Our result comparing with GL inpainting~\cite{IizukaSIGGRAPH2017}. (a) \& (d) The input image with missing hole. (b) \& (d) Inpainting result given by GL inpainting~\cite{IizukaSIGGRAPH2017}. (c) \& (f) Final inpainting result using our approach. The size of images are 512x512.}
\label{fig:teaser}
\end{figure}
\noindent
The problem of generating photo-realistic images from sampled noise or conditioning on other inputs such as images, texts or labels has been heavily investigated. In spite of recent progress of deep generative models such as PixelCNN~\cite{van2016conditional}, VAE~\cite{kingma2013auto} and GANs~\cite{goodfellow2014generative}, generating high-resolution images remains a difficult task. This is mainly because modeling the distribution of pixels is difficult and the trained models easily introduce blurry components and artifacts when the dimensionality becomes high. Several approaches have been proposed to alleviate the problem, usually by leveraging multi-scale training~\cite{zhang2016stackgan,denton2015deep} or incorporating prior information~\cite{nguyen2016plug}.

In addition to the general image synthesis problem, the task of image inpainting can be described as: given an incomplete image as input, how do we fill in the missing parts with semantically and visually plausible contents. We are interested in this problem for several reasons. First, it is a well-motivated task for a common scenario where we may want to remove unwanted objects from pictures or restore damaged photographs. Second, while purely unsupervised learning may be challenging for large inputs, we show in this work that the problem becomes more constrained and tractable when we train in a multi-stage self-supervised manner and leverage the high-frequency information in the known region. 

Context-encoder~\cite{pathak2016context} is one of the first works that apply deep neural networks for image inpainting. It trains a deep generative model that maps an incomplete image to a complete image using reconstruction loss and adversarial loss. While adversarial loss significantly improves the inpainting quality, the results are still quite blurry and contain notable artifacts. In addition, we found it fails to produce reasonable results for larger inputs like 512x512 images, showing it is unable generalize to high-resolution inpainting task. More recently,~\cite{IizukaSIGGRAPH2017} improved the result by using dilated convolution and an additional local discriminator. However it is still limited to relatively small images and holes due to the spatial support of the model. 

Yang \emph{et al.}~\cite{Yang_2017_CVPR} proposes to use style transfer for image inpainting. More specifically, it initializes the hole with the output of context-encoder, and then improves the texture by using style transfer techniques~\cite{li2016combining} to propagate the high-frequency textures from the boundary to the hole. It shows that matching the neural features not only transfers artistic styles, but can also synthesize real-world images. The approach is optimization-based and applicable to images of arbitrary sizes. However, the computation is costly and it takes long time to inpaint a large image.   

Our approach overcomes the limitation of the aforementioned methods. Being similar to~\cite{Yang_2017_CVPR}, we decouple the inpainting process into two stages: inference and translation. In the inference stage, we train an \textit{Image2Feature} network that initializes the hole with coarse prediction and extract its features. The prediction is blurry but contains high-level structure information in the hole. In the translation stage, we train a \textit{Feature2Image} network that transforms the feature back into a complete image. It refines the contents in the hole and outputs a complete image with sharp and realistic texture. Its main difference with~\cite{Yang_2017_CVPR} is that, instead of relying on optimization, we model texture refinement as a learning problem. Both networks can be trained end-to-end and, with the trained models, the inference can be done in a single forward pass, which is much faster than iterative optimizations.

To ease the difficulty of training the Feature2Image network, we design a ``patch-swap'' layer that propagates the high-frequency texture details from the boundary to the hole. The patch-swap layer takes the feature map as input, and replaces each neural patch inside the hole with the most similar patch on the boundary. We then use the new feature map as the input to the Feature2Image network. By re-using the neural patches on the boundary, the feature map contains sufficient details, making the high-resolution image reconstruction feasible. 

We note that by dividing the training into two stages of Image2Feature and Feature2Image greatly reduces the dimensionality of possible mappings between input and output. Injecting prior knowledge with patch-swap further guides the training process such that it is easier to find the optimal transformation. When being compared with the GL inpainting~\cite{IizukaSIGGRAPH2017}, we generate sharper and better inpainting results at size 256x256. Our approach also scales to higher resolution (i.e. 512x512), which GL inpainting fails to handle. As compared with neural inpainting~\cite{Yang_2017_CVPR}, our results have comparable or better visual quality in most examples. In particular, our synthesized contents blends with the boundary more seamlessly. Our approach is also much faster.

The main contributions of this paper are:
(1) We design a learning-based inpainting system that is able to synthesize missing parts in a high-resolution image with high-quality contents and textures.
(2) We propose a novel and robust training scheme that addresses the issue of feature manipulation and avoids under-fitting. 
(3) We show that our trained model can achieve performance comparable with state-of-the-art and generalize to other tasks like style transfer.


\section{Related Work}

Image generation with generative adversarial networks (GANs) has gained remarkable progress recently. The vanilla GANs~\cite{goodfellow2014generative} has shown promising performance to generate sharp images, but training instability makes it hard to scale to higher resolution images. Several techniques have been proposed to stabilize the training process, including DCGAN~\cite{radford2015unsupervised}, energy-based GAN~\cite{zhao2016energy}, Wasserstein GAN (WGAN)~\cite{salimans2016improved,arjovsky2017wasserstein}, WGAN-GP~\cite{gulrajani2017improved}, BEGAN~\cite{berthelot2017began}, LSGAN~\cite{mao2016least} and the more recent Progressive GANs~\cite{karras2017progressive}. A more relevant task to inpainting is conditional image generation. For example, Pix2Pix~\cite{isola2016image}, Pix2Pix HD~\cite{wang2017high} and CycleGAN~\cite{zhu2017unpaired} translate images across different domains using paired or unpaired data. Using deep neural network for image inpainting has also been studied in~\cite{yeh2017semantic,pathak2016context,Yang_2017_CVPR,wang2017shape,IizukaSIGGRAPH2017}.   

Our patch-swap can be related to recent works in neural style transfer. Gatys \emph{et al.}~\cite{gatys2015neural} first formulates style transfer as an optimization problem that combines texture synthesis with content reconstruction. As an alternative,~\cite{elad2017style,frigo2016split,van2016conditional} use neural-patch based similarity matching between the content and style images for style transfer. Li and Wand~\cite{li2016combining} optimize the output image such that each of the neural patch matches with a similar neural patch in the style image. This enables arbitrary style transfer at the cost of expensive computation.~\cite{chen2016fast} proposes a fast approximation to~\cite{li2016combining} where it constructs the feature map directly and uses an inverse network to synthesize the image in feed-forward manner. 

Traditional non-neural inpainting algorithms~\cite{barnes2009patchmatch,barnes2010generalized} mostly work on the image space. While they share similar ideas of patch matching and propagation, they are usually agnostic to high-level semantic and structural information. 
\section{Methodology}\label{sec:approach}
\begin{figure*}[t]
	\centering
	\includegraphics[width=.98\linewidth]{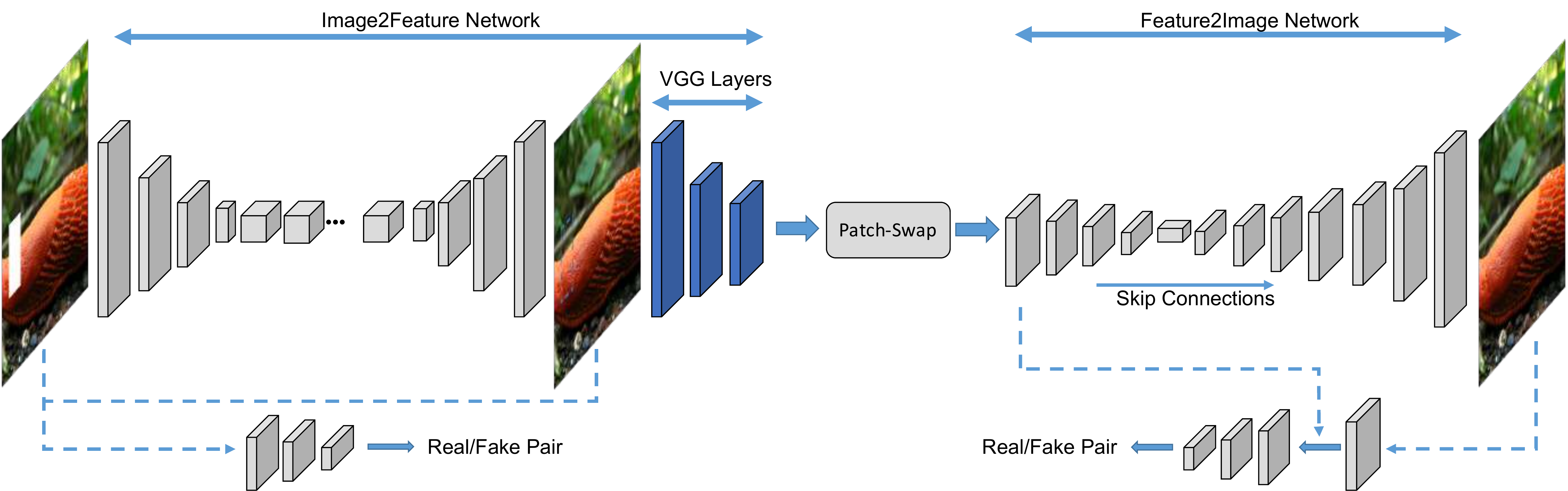}
	\caption{Overview of our network architecture. We use Image2Feature network as coarse inferrence and use VGG network to extract a feature map. Then patch-swap matches neural patches from boundary to the hole. Finally the Feature2Image network translates to a complete, high-resolution image.} 
\end{figure*}
\label{fig:framework}

\subsection{Problem Description}

We formalize the task of image inpainting as follows: suppose we are given an incomplete input image $I_0$, with $R$ and $\bar{R}$ representing the missing region (the hole) and the known region (the boundary) respectively. We would like to fill in $R$ with plausible contents $I_R$ and combine it with $I_0$ as a new, complete image $I$. Evaluating the quality of inpainting is mostly subject to human perception but ideally, $I_R$ should meet the following criteria: 1. It has sharp and realistic-looking textures; 2. It contains meaningful content and is coherent with $I_{\bar{R}}$ and 3. It looks like what appears in the ground truth image $I_{gt}$ (if available). In our context, $R$ can be either a single hole or multiple holes. It may also come with arbitrary shape, placed on a random location of the image.

\subsection{System Overview}
Our system divides the image inpainting tasks into three steps:\\
\textbf{Inference:} We use an Image2Feature network to fill an incomplete image with coarse contents as inference and extract a feature map from the inpainted image. \\
\textbf{Matching:} We use patch-swap on the feature map to match the neural patches from the high-resolution boundary to the hole with coarse inference.\\
\textbf{Translation:} We use a Feature2Image network to translate the feature map to a complete image.\\
The entire pipeline is illustrated in Fig.~\ref{fig:framework}.

\subsection{Training}\label{sec:training}
We introduce separate steps of training the Image2Feature and Feature2Image network. For illustration purpose, we assume the size of $I_0$ is 256x256x3 and the hole $R$ has size 128x128.

\subsubsection*{Inference: Training Image2Feature Network}
The goal of the Image2Feature network is to fill in the hole with coarse prediction. During training, the input to the Image2Feature translation network is the 256x256x3 incomplete image $I_0$ and the output is a feature map $F_1$ of size 64x64x256. The network consists of an FCN-based module $G_1$, which consists of a down-sampling front end, multiple intermediate residual blocks and an up-sampling back end. $G_1$ is followed by the initial layers of the 19-layer VGG network~\cite{simonyan2014very}. Here we use the filter pyramid of the VGG network as a higher-level representation of images similar to~\cite{gatys2015neural}. At first, $I_0$ is given as input to $G_1$ which produces a coarse prediction $I_1^R$ of size 128x128. $I_1^R$ is then embedded into $R$ forming a complete image $I_1$, which again passes through the VGG19 network to get the activation of \emph{relu3\_1} as $F_1$. $F_1$ has size 64x64x256. We also use an additional PatchGAN discriminator $D_1$ to facilitate adversarial training, which takes a pair of images as input and outputs a vector of true/fake probabilities.

For $G_1$, the down-sampling front-end consists of three convolutional layers, and each layer has stride 2. The intermediate part has 9 residual blocks stacked together. The up-sampling back-end is the reverse of the front-end and consists of three transposed convolution with stride 2. Every convolutional layer is followed by batch normalization~\cite{ioffe2015batch} and ReLu activation, except for the last layer which outputs the image. We also use dilated convolution in all residual blocks. Similar architecture has been used in~\cite{wang2017high} for image synthesis and~\cite{IizukaSIGGRAPH2017} for inpainting. Different from~\cite{wang2017high}, we use dilated layer to increase the size of receptive field. Comparing with~\cite{IizukaSIGGRAPH2017}, our receptive field is also larger given we have more down-sampling blocks and more dilated layers in residual blocks.

During training, the overall loss function is defined as:
\begin{eqnarray}
L_{G_1} =  \lambda_{1}L_{perceptual}+\lambda_{2}L_{adv}.
\end{eqnarray}
The first term is the perceptual loss, which is shown to correspond better with human perception of similarity~\cite{zhang2018unreasonable} and has been widely used in many tasks~\cite{gatys2016image,johnson2016perceptual,dosovitskiy2016generating,chen2016fast}:
\begin{eqnarray}
L_{perceptual}(F, I_gt) &=& \parallel \mathcal{M}_F\circ(F_1-vgg(I_{gt}))\parallel_1.
\end{eqnarray}
\label{eqn:rec}
Here $\mathcal{M}_F$ are the weighted masks yielding the loss to be computed only on the hole of the feature map. We also assign higher weight to the overlapping pixels between the hole and the boundary to ensure the composite is coherent. The weights of VGG19 network are loaded from the ImageNet pre-trained model and are fixed during training. 

The adversarial loss is based on Generative Adversarial Networks (GANs) and is defined as:
\begin{eqnarray}
L_{adv} = \max_{D_1}E[\log(D_1(I_0, I_{gt}))+\log(1-D_1(I_0, I_1))].
\end{eqnarray}
We use a pair of images as input to the discriminator. Under the setting of adversarial training, the real pair is the incomplete image $I_0$ and the original image $I_{gt}$, while the fake pair is $I_0$ and the prediction $I_1$.

To align the absolute value of each loss, we set the weight $\lambda_{1}=10$ and $\lambda_{2}=1$ respectively. We use Adam optimizer for training. The learning rate is set as $lr_{G}=2\mathrm{e}{-3}$ and $lr_{D}=2\mathrm{e}{-4}$ and the momentum is set to 0.5.

\subsubsection*{Match: Patch-swap Operation}
\begin{figure}
\centering
\small
\includegraphics[width=.48\textwidth]{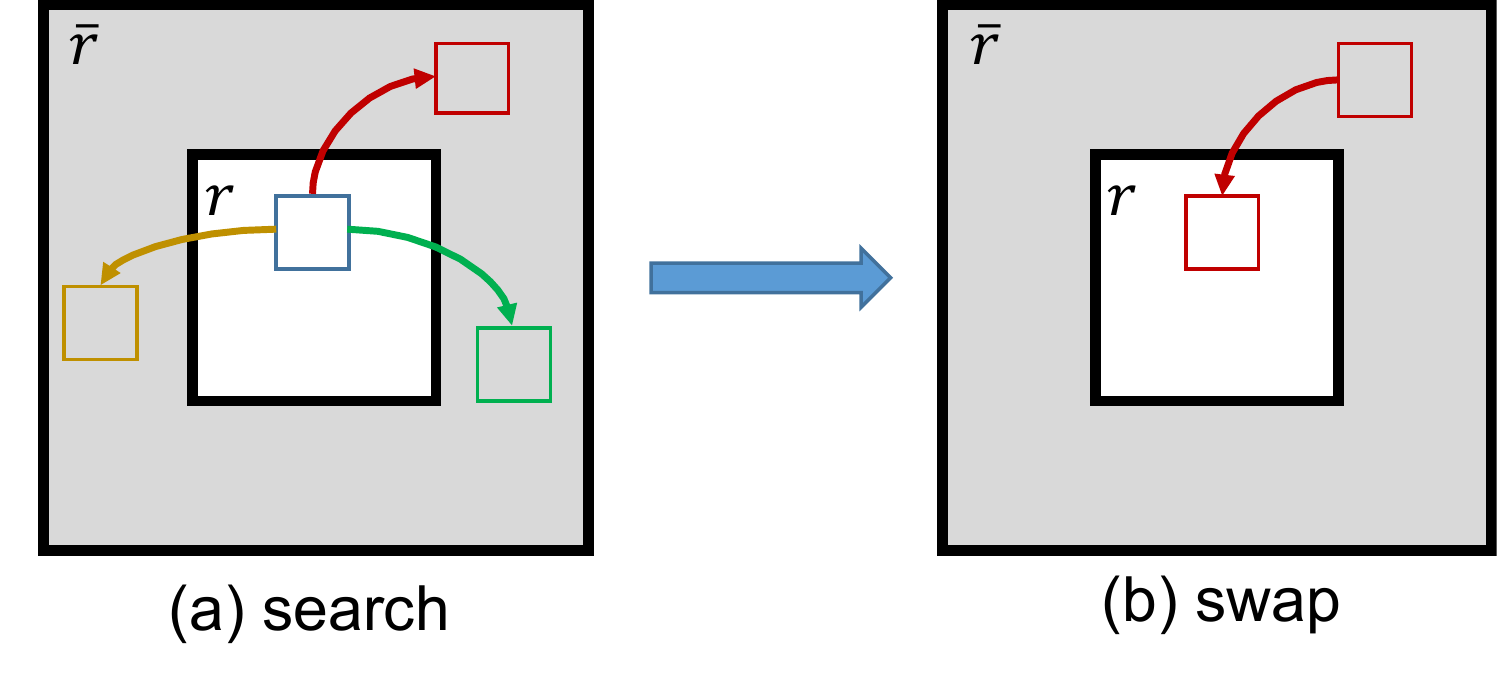}
\caption{Illustration of patch-swap operation. Each neural patch in the hole $r$ searches for the most similar neural patch on the boundary $\bar{r}$, and then swaps with that patch.}
\label{fig:patchswap}
\end{figure}

Patch-swap is an operation which transforms $F_1$ into a new feature map $F'_1$. The idea is that the prediction $I_1^R$ is blurry, lacking many of the high-frequency details. Intuitively, we would like to propagate the textures from $I_1^{\bar{R}}$ onto $I_1^R$ but still preserves the high-level information of $I_1^R$. Instead of operating on $I_1$ directly, we use $F_1$ as a surrogate for texture propagation. Similarly, we use $r$ and $\bar{r}$ to denote the region on $F_1$ corresponding to $R$ and $\bar{R}$ on $I_1$. For each 3x3 neural patch $p_i (i=1,2,...,N)$ of $F_1$ overlapping with $r$, we find the closest-matching neural patch in $\bar{r}$ based on the following cross-correlation metric:
\begin{eqnarray}
d(p,p') =\frac{<p,p'>}{\parallel p\parallel\cdot \parallel p'\parallel}
\end{eqnarray}
Suppose the closest-matching patch of $p_i$ is $q_i$, we then replace $p_i$ with $q_i$. After each patch in $r$ is swapped with its most similar patch in $\bar{r}$, overlapping patches are averaged and the output is a new feature map $F'_1$. We illustrate the process in Fig.~\ref{fig:patchswap}.

Measuring the cross-correlations for all the neural patch pairs between the hole and boundary is computationally expensive. To address this issue, we follow similar implementation in~\cite{chen2016fast} and speed up the computation using paralleled convolution. We summarize the algorithm as following steps. First, we normalize and stack the neural patches on $\bar{r}$ and view the stacked vector as a convolution filter. Next, we apply the convolution filter on $r$. The result is that at each location of $r$ we get a vector of values which is the cross-correlation between the neural patch centered at that location and all patches in $\bar{r}$. Finally, we replace the patch in $r$ with the patch in $\bar{r}$ of maximum cross-correlation. Since the whole process can be parallelized, the amount of time is significantly reduced. In practice, it only takes about 0.1 seconds to process a 64x64x256 feature map.

\subsubsection*{Translate: Training Feature2Image Translation Network}
The goal of the Feature2Image network is to learn a mapping from the swapped feature map to a complete and sharp image. It has a U-Net style generator $G_2$ which is similar to $G_1$, except the number of hidden layers are different. The input to $G_2$ is a feature map of size 64x64x256. The generator has seven convolution blocks and eight deconvolution blocks, and the first six deconvolutional layers are connected with the convolutional layers using skip connection. The output is a complete 256x256x3 image. It also consists of a Patch-GAN based discriminator $D_2$ for adversarial training. However different from the Image2Feature network which takes a pair of images as input, the input to $D_2$ is a pair of image and feature map.

A straightforward training paradigm is to use the output of the Image2Feature network $F_1$ as input to the patch-swap layer, and then use the swapped feature $F'_1$ to train the Feature2Image model. In this way, the feature map is derived from the coarse prediction $I_1$ and the whole system can be trained end-to-end. However, in practice, we found that this leads to poor-quality reconstruction $I$ with notable noise and artifacts (Sec.~\ref{sec:exp}). We further observed that using the ground truth as training input gives rise to results of significantly improved visual quality. That is, we use the feature map $F_{gt}=\mbox{vgg}(I_{gt})$ as input to the patch-swap layer, and then use the swapped feature $F'_{gt}=\mbox{patch\_swap}(F_{gt})$ to train the Feature2Image model. Since $I_{gt}$ is not accessible at test time, we still use $F'_1=\mbox{patch\_swap}(F_1)$ as input for inference. Note that now the Feature2Image model trains and tests with different types of input, which is not a usual practice to train a machine learning model.  

Here we provide some intuition for this phenomenon. Essentially by training the Feature2Image network, we are learning a mapping from the feature space to the image space. Since $F_1$ is the output of the Image2Feature network, it inherently contains a significant amount of noise and ambiguity. Therefore the feature space made up of $F'_1$ has much higher dimensionality than the feature space made up of $F'_{gt}$. The outcome is that the model easily under-fits $F'_1$, making it difficult to learn a good mapping. Alternatively, by using $F'_{gt}$, we are selecting a clean, compact subset of features such that the space of mapping is much smaller, making it easier to learn. Our experiment also shows that the model trained with ground truth generalizes well to noisy input $F'_1$ at test time. Similar to~\cite{zheng2016improving}, we can further improve the robustness by sampling from both the ground truth and Image2Feature prediction.  

The overall loss for the Feature2Image translation network is defined as:
\begin{eqnarray}
L_{G_2} = \lambda_{1}L_{perceptual}+\lambda_{2}L_{adv}.
\end{eqnarray} 
The reconstruction loss is defined on the entire image between the final output $I$ and the ground truth $I_{gt}$:
\begin{eqnarray}
L_{perceptual}(I, I_{gt}) =\parallel vgg(I)-vgg(I_{gt})\parallel_2. 
\end{eqnarray} 

The adversarial loss is given by the discriminator $D_2$ and is defined as:
\begin{eqnarray}
L_{adv} = \max_{D_2}E[\log(D_2(F'_{gt}, I_{gt}))+\log(1-D_2(F'_{gt}, I))].
\end{eqnarray}
The real and fake pair for adversarial training are ($F'_{gt},I_{gt}$) and ($F'_{gt},I$). 

When training the Feature2Image network we set $\lambda_{1}=10$ and $\lambda_{2}=1$. For the learning rate, we set $lr_{G}=2\mathrm{e}{-4}$ and $lr_{D}=2\mathrm{e}{-4}$. Same as the Image2Feature network, the momentum is set to 0.5.

\subsection{Multi-scale Inference}

Given the trained models, inference is straight-forward and can be done in a single forward pass. The input $I_0$ successively passes through the Image2Feature network to get $I_1$ and $F_1=\mbox{vgg}(I_1)$, then the patch-swap layer ($F'_1$), and then finally the Feature2Image network ($I$). We then use the center of $I$ and blend with $I_0$ as the output.

\begin{figure}
\centering
\small
\includegraphics[width=.72\textwidth]{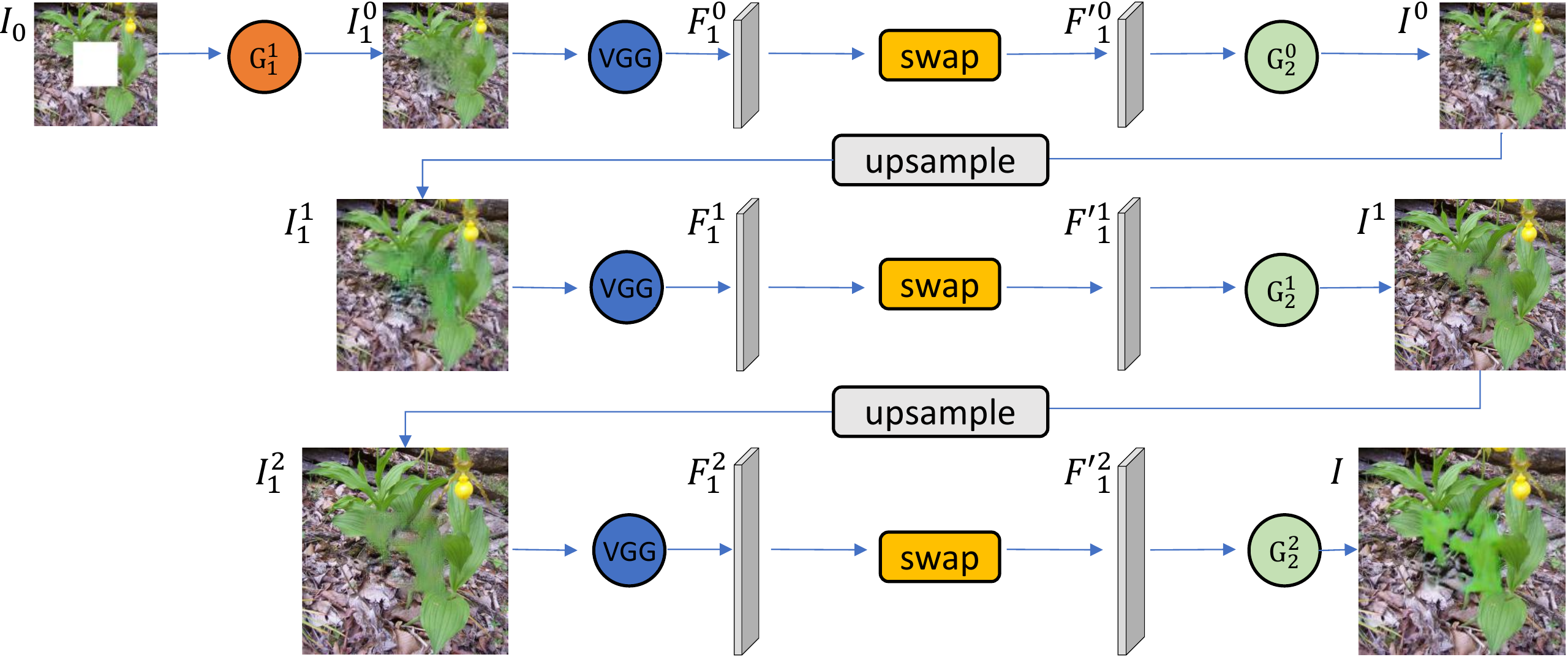}
\caption{Multi-scale inference.}
\label{fig:multiscale}
\end{figure}

Our framework can be easily adapted to multi-scale. The key is that we directly upsample the output of the lower scale as the input to the Feature2Image network of the next scale (after using VGG network to extract features and apply patch-swap). In this way, we will only need the Image2Feature network at the smallest scale $s_0$ to get $I^0_1$ and $F^0_1$. At higher scales $s_i (i>0)$ we simply set $I^{s_i}_1=\mbox{upsample}(I^{s_{i-1}})$ and let $F^{s_i}_1=\mbox{vgg}(I^{s_i}_1)$ (Fig.~\ref{fig:multiscale}). Training Image2Feature network can be challenging at high resolution. However by using the multi-scale approach we are able to initialize from lower scales instead, allowing us to handle large inputs effectively. We use multi-scale inference on all our experiments.


\section{Experiments}\label{sec:exp}

\subsection{Experiment Setup}
We separately train and test on two public datasets: COCO~\cite{lin2014microsoft} and ImageNet CLS-LOC~\cite{russakovsky2015imagenet}. The number of training images in each dataset are: 118,287 for COCO and 1,281,167 for ImageNet CLS-LOC. We compare with content aware fill (CAF)~\cite{barnes2009patchmatch}, context encoder (CE)~\cite{pathak2016context}, neural patch synthesis (NPS)~\cite{Yang_2017_CVPR} and global local inpainting (GLI)~\cite{IizukaSIGGRAPH2017}. For CE, NPS, and GLI, we used the public available trained model. CE and NPS are trained to handle fixed holes, while GLI and CAF can handle arbitrary holes. To fairly evaluate, we experimented on both settings of fixed hole and random hole. For fixed hole, we compare with CAF, CE, NPS, and GLI on image size 512x512 from ImageNet test set. The hole is set to be 224x224 located at the image center. For random hole, we compare with CAF and GLI, using COCO test images resized to 256x256. In the case of random hole, the hole size ranges from 32 to 128 and is placed anywhere on the image. We observed that for small holes on 256x256 images, using patch-swap and Feature2Image network to refine is optional as our Image2Feature network already generates satisfying results most of the time. While for 512x512 images, it is necessary to apply multi-scale inpainting, starting from size 256x256. To address both sizes and to apply multi-scale, we train the Image2Feature network at 256x256 and train the Feature2Image network at both 256x256 and 512x512. During training, we use early stopping, meaning we terminate the training when the loss on the held-out validation set converges. On our NVIDIA GeForce GTX 1080Ti GPU, training typically takes one day to finish for each model, and test time is around 400ms for a 512x512 image.

\subsection{Results}
\noindent\textbf{Quantitative Comparison}
Table~\ref{table:numerical} shows numerical comparison result between our approach, CE~\cite{pathak2016context}, GLI~\cite{IizukaSIGGRAPH2017} and NPS~\cite{Yang_2017_CVPR}. We adopt three quality measurements: mean $\ell_1$ error, SSIM, and inception score~\cite{salimans2016improved}. Since context encoder only inpaints 128x128 images and we failed to train the model for larger inputs, we directly use the 128x128 results and bi-linearly upsample them to 512x512. Here we also compute the SSIM over the hole area only. We see that although our mean $\ell_1$ error is higher, we achieve the best SSIM and inception score among all the methods, showing our results are closer to the ground truth by human perception. Besides, mean $\ell_1$ error is not an optimal measure for inpainting, as it favors averaged colors and blurry results and does not directly account for the end goal of perceptual quality.   

\begin{table}
\caption{Numerical comparison on 200 test images of ImageNet.}
\begin{center}

\resizebox{.75\textwidth}{!}{%
{\tiny
  \begin{tabular}{ l  c  c  c}
    \hline
    \textbf{Method} & \textbf{Mean $\ell_1$ Error} &  \textbf{SSIM} & \textbf{Inception Score}  \\ \hline
    \emph{CE~\cite{pathak2016context}} & 15.46\% & 0.45 & 9.80\\ \hline
    \emph{NPS~\cite{Yang_2017_CVPR}} & \textbf{15.13\%} & 0.52 & 10.85 \\ \hline
    \emph{GLI~\cite{IizukaSIGGRAPH2017}} & 15.81\% & 0.55 & 11.18 \\ \hline
    \emph{our approach} & 15.61\% & \textbf{0.56} & \textbf{11.36} \\ \hline
    \hline
  \end{tabular}}
  }
  \end{center}
  \label{table:numerical}
\end{table}

\begin{figure*}
\centering
\small
\setlength{\tabcolsep}{1pt}
\begin{tabular}{cccccc}
 \includegraphics[width=.16\textwidth]{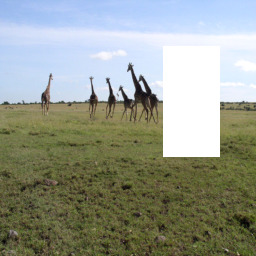}&
 \includegraphics[width=.16\textwidth]{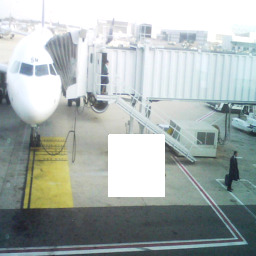}&
 \includegraphics[width=.16\textwidth]{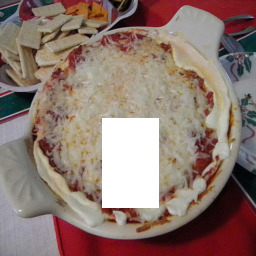}&
 \includegraphics[width=.16\textwidth]{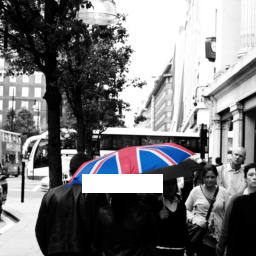}&
 \includegraphics[width=.16\textwidth]{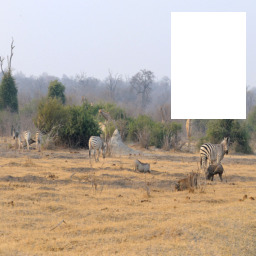}&
 \includegraphics[width=.16\textwidth]{figures/result_random_hole/figures/000000502136_input_image.jpg}\\
 \includegraphics[width=.16\textwidth]{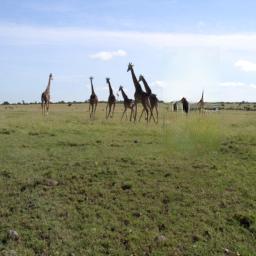}&
 \includegraphics[width=.16\textwidth]{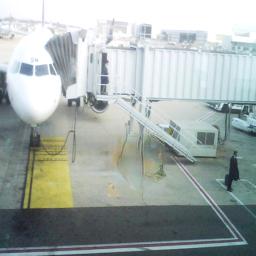}&
 \includegraphics[width=.16\textwidth]{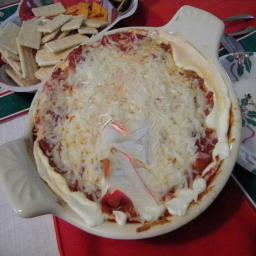}&
 \includegraphics[width=.16\textwidth]{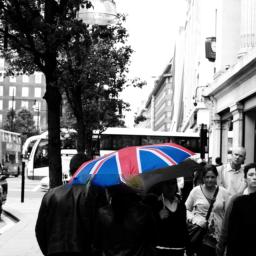}&
 \includegraphics[width=.16\textwidth]{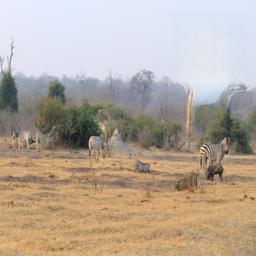}&
 \includegraphics[width=.16\textwidth]{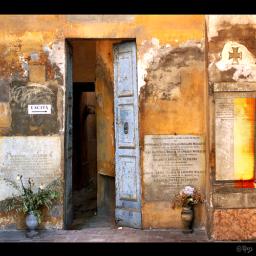}\\
 \includegraphics[width=.16\textwidth]{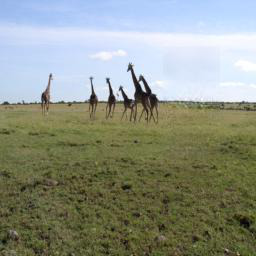}&
 \includegraphics[width=.16\textwidth]{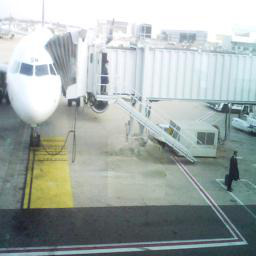}&
 \includegraphics[width=.16\textwidth]{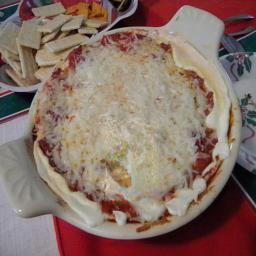}&
 \includegraphics[width=.16\textwidth]{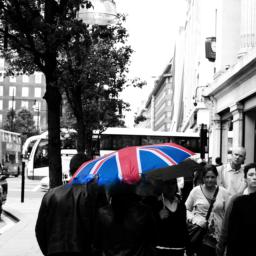}&
 \includegraphics[width=.16\textwidth]{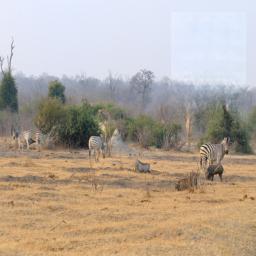}&
 \includegraphics[width=.16\textwidth]{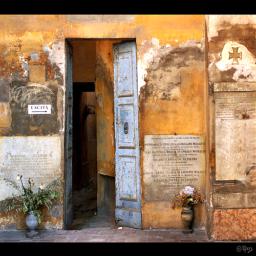}\\
\end{tabular}
\caption{Visual comparisons of ImageNet results with random hole. Each example from top to bottom: input image, GLI~\cite{IizukaSIGGRAPH2017}, our result. All images have size $256\times 256$.}
\label{fig:randomhole}
\end{figure*}

\noindent\textbf{Visual Result}
Fig.~\ref{fig:randomhole} shows our comparison with GLI [1] in random hole cases. We can see that our method could handle multiple situations better, such as object removal, object completion and texture generation, while GLI’s results are noisier and less coherent. From Fig.~\ref{fig:comparison}, we could also find that our results are better than GLI most of the time for large holes. This shows that directly training a network for large hole inpainting is difficult, and it is where our ``patch-swap'' can be most helpful. In addition, our results have significantly fewer artifacts than GLI. Comparing with CAF, we can better predict the global structure and fill in contents more coherent with the surrounding context. Comparing with CE, we can handle much larger images and the synthesized contents are much sharper. Comparing with NPS whose results mostly depend on CE, we have similar or better quality most of the time, and our algorithm also runs much faster. Meanwhile, our final results improve over the intermediate output of Image2Feature. This demonstrates that using patch-swap and Feature2Image transformation is beneficial and necessary.

\noindent\textbf{User Study} To better evaluate and compare with other methods, we randomly select 400 images from the COCO test set and randomly distribute these images to 20 users. Each user is given 20 images with holes together with the inpainting results of NPS, GLI, and ours. Each of them is asked to rank the results in non-increasing order (meaning they can say two results have similar quality). We collected 399 valid votes in total found our results are ranked best most of the time: in 75.9\% of the rankings our result receives highest score. In particular, our results are overwhelmingly better than GLI, receiving higher score 91.2\% of the time. This is largely because GLI does not handle large holes well. Our results are also comparable with NPS, ranking higher or the same 86.2\% of the time. 

\subsection{Analysis}\label{sec:analysis}

\begin{figure*}[t]
  \center
\setlength\tabcolsep{1pt}
\begin{tabular}{ccccccc}
 \includegraphics[width=.14\textwidth]{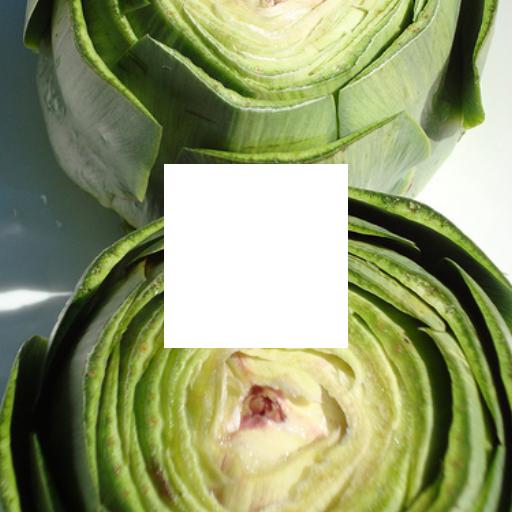}&
 \includegraphics[width=.14\textwidth]{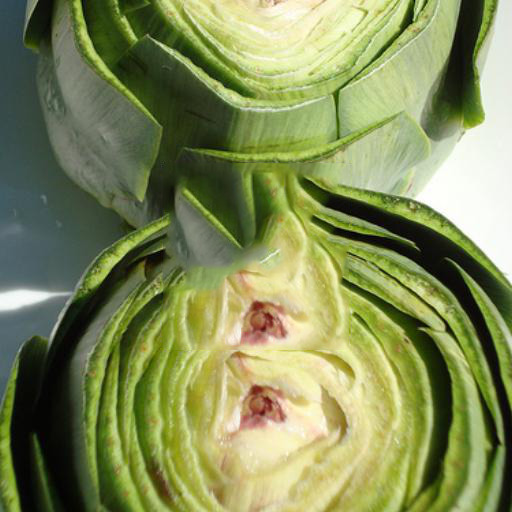}&
 \includegraphics[width=.14\textwidth]{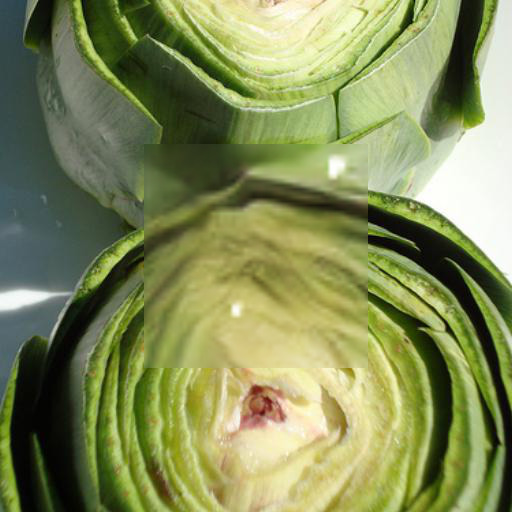}&
 \includegraphics[width=.14\textwidth]{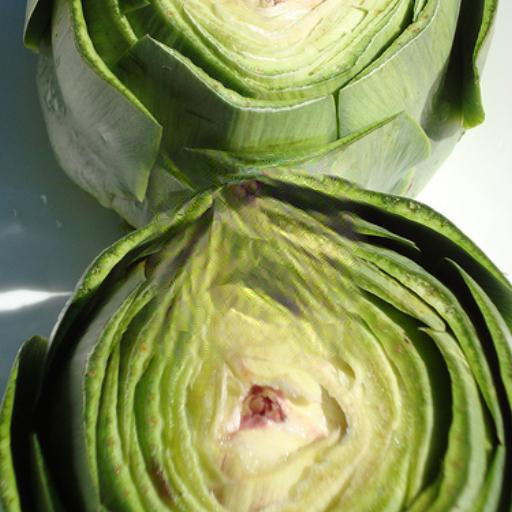}&
 \includegraphics[width=.14\textwidth]{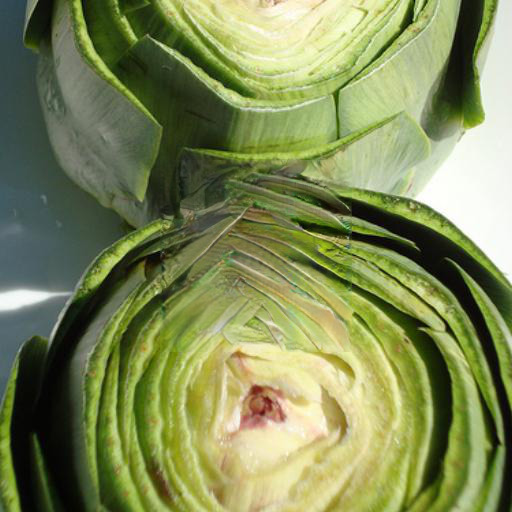}&
 \includegraphics[width=.14\textwidth]{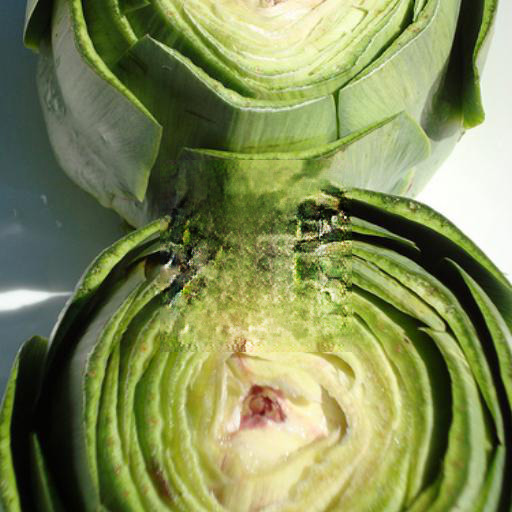}&
 \includegraphics[width=.14\textwidth]{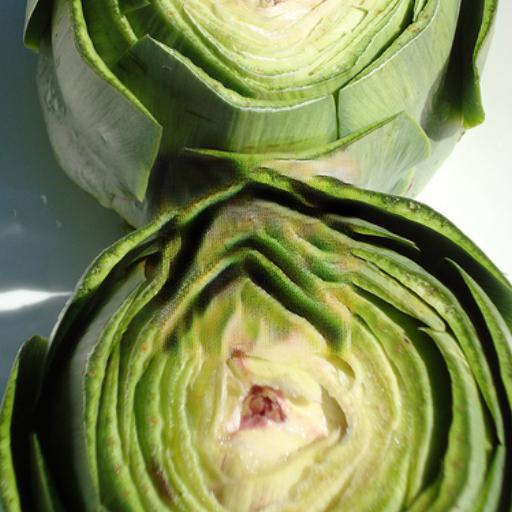}\\
 \includegraphics[width=.14\textwidth]{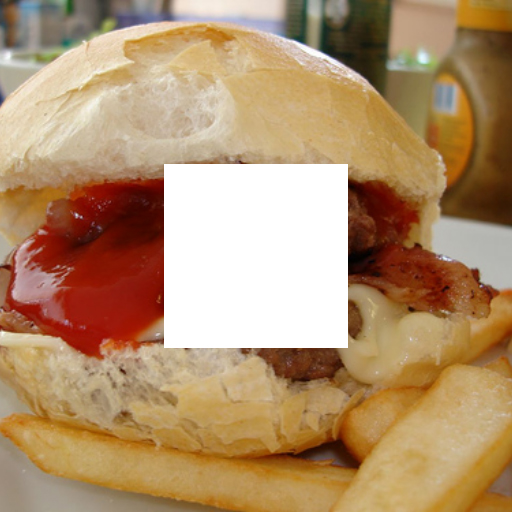}&
 \includegraphics[width=.14\textwidth]{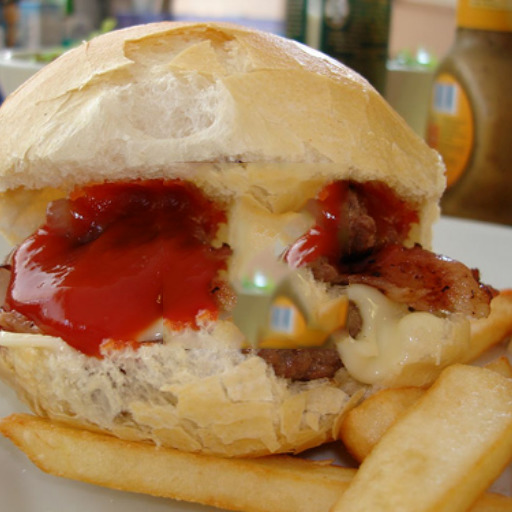}&
 \includegraphics[width=.14\textwidth]{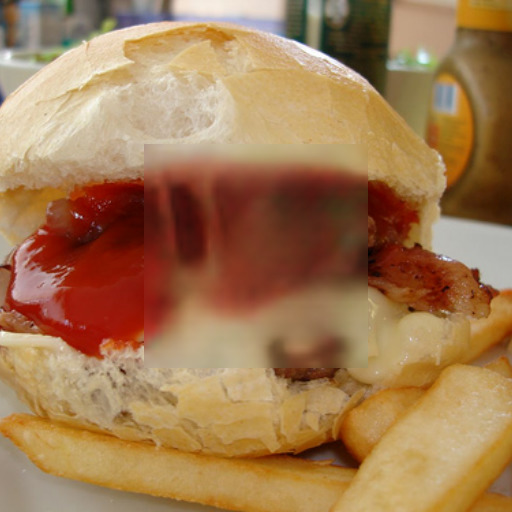}&
 \includegraphics[width=.14\textwidth]{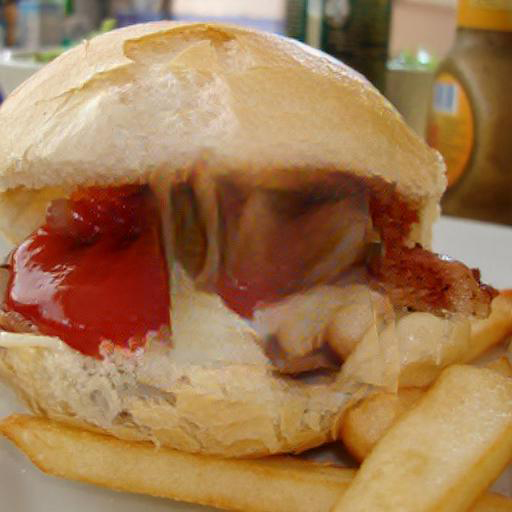}&
 \includegraphics[width=.14\textwidth]{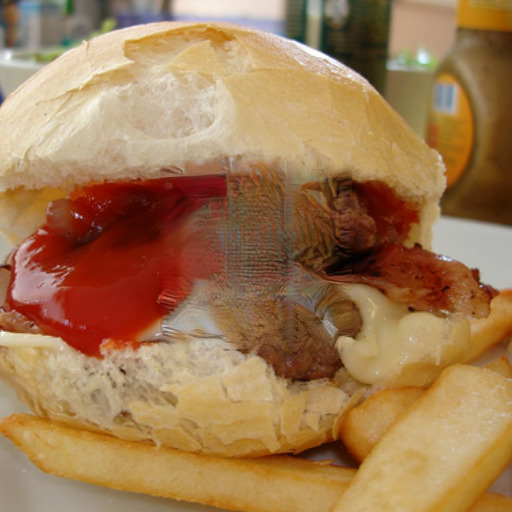}&
 \includegraphics[width=.14\textwidth]{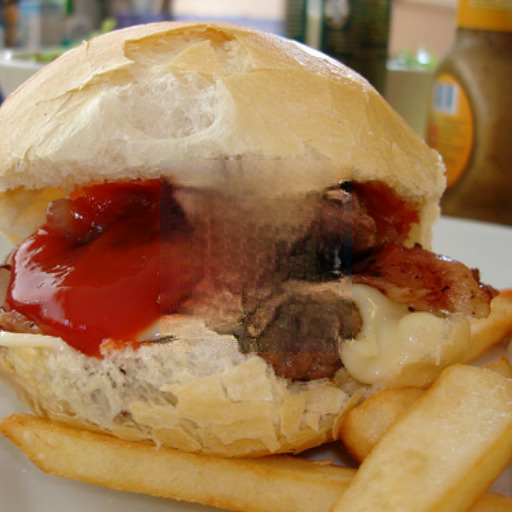}&
 \includegraphics[width=.14\textwidth]{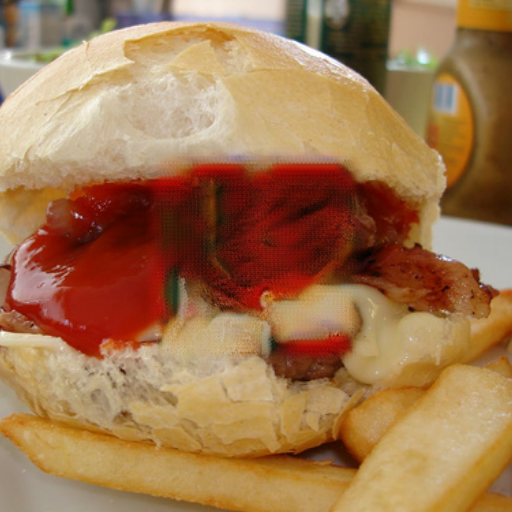}\\
 \includegraphics[width=.14\textwidth]{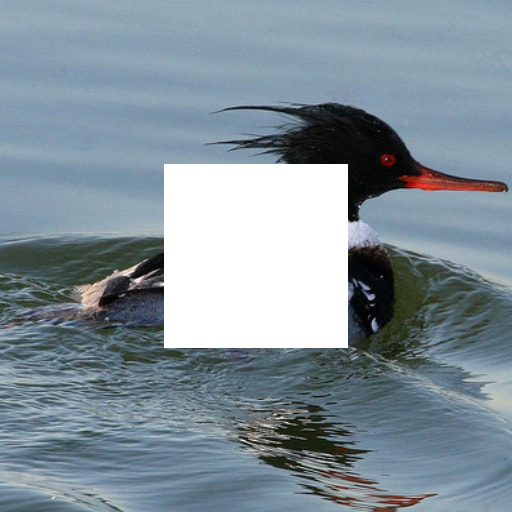}&
 \includegraphics[width=.14\textwidth]{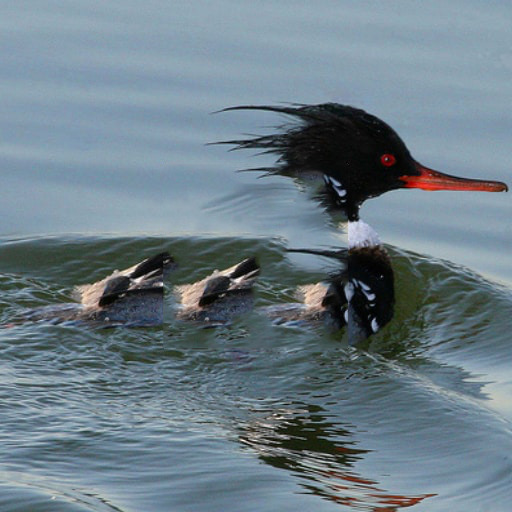}&
 \includegraphics[width=.14\textwidth]{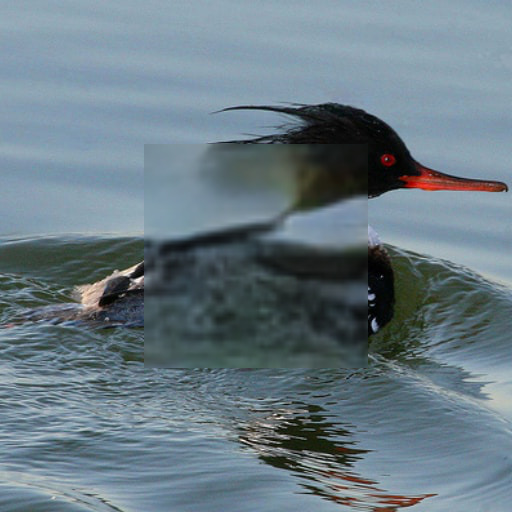}&
 \includegraphics[width=.14\textwidth]{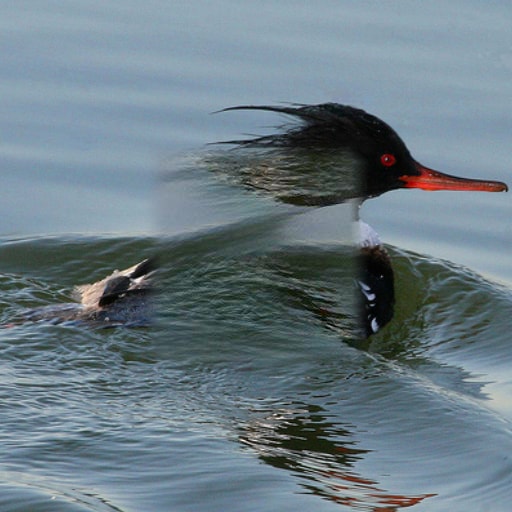}&
 \includegraphics[width=.14\textwidth]{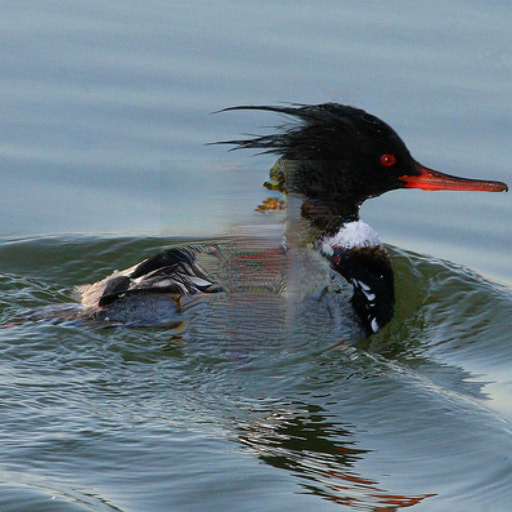}&
 \includegraphics[width=.14\textwidth]{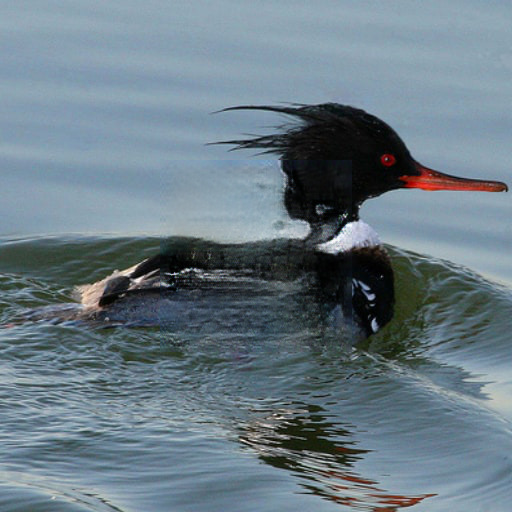}&
 \includegraphics[width=.14\textwidth]{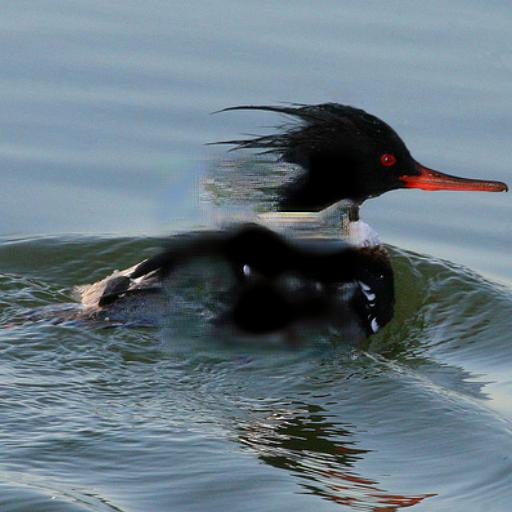}\\
 \includegraphics[width=.14\textwidth]{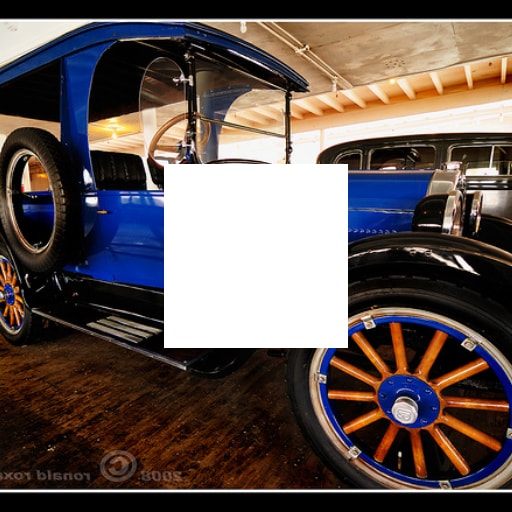}&
 \includegraphics[width=.14\textwidth]{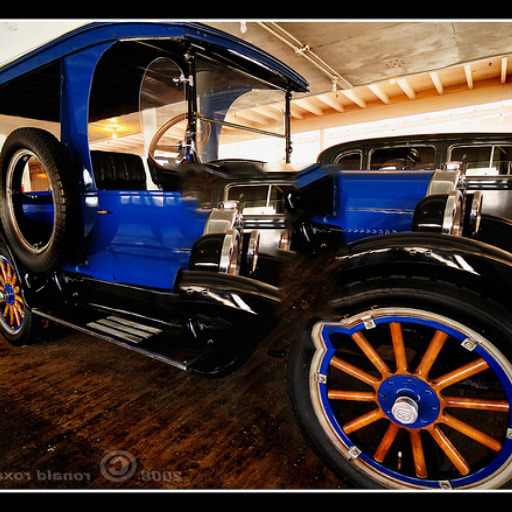}&
 \includegraphics[width=.14\textwidth]{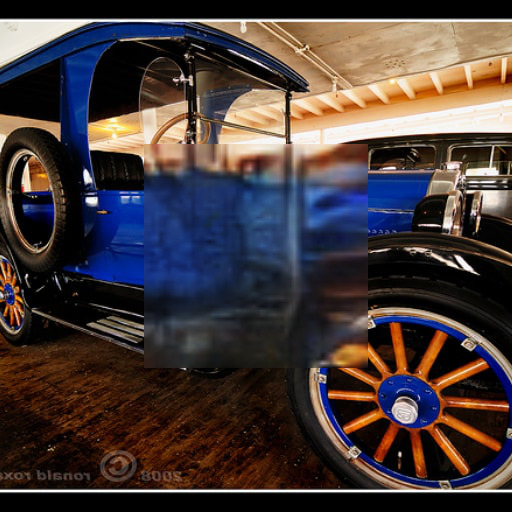}&
 \includegraphics[width=.14\textwidth]{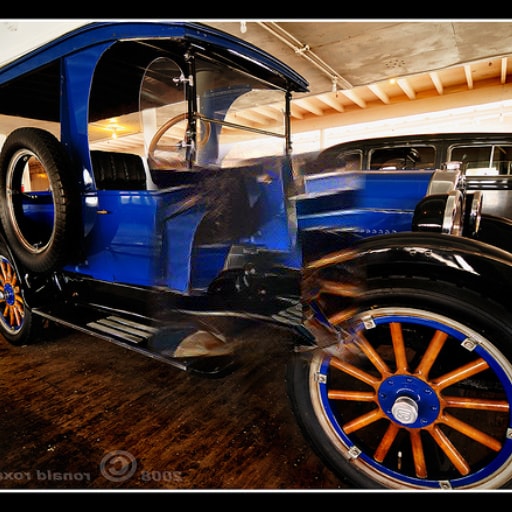}&
 \includegraphics[width=.14\textwidth]{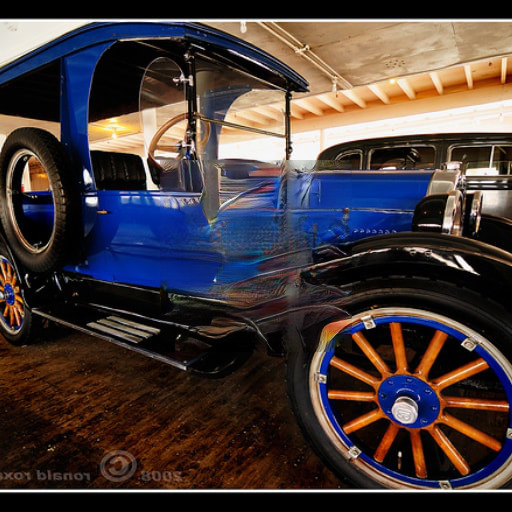}&
 \includegraphics[width=.14\textwidth]{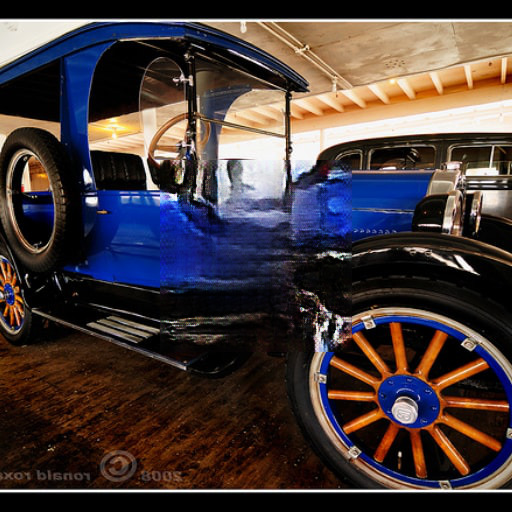}&
 \includegraphics[width=.14\textwidth]{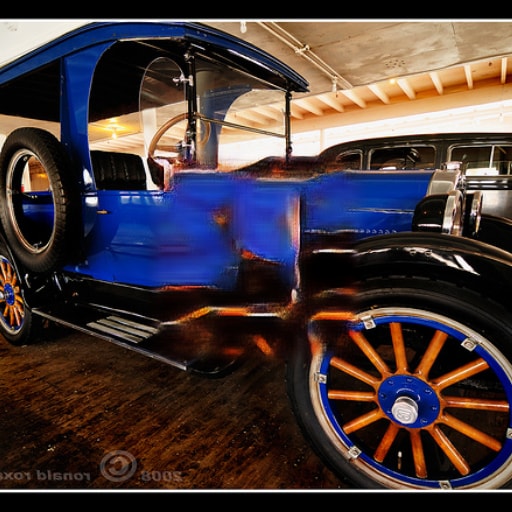}\\
   \includegraphics[width=.14\textwidth]{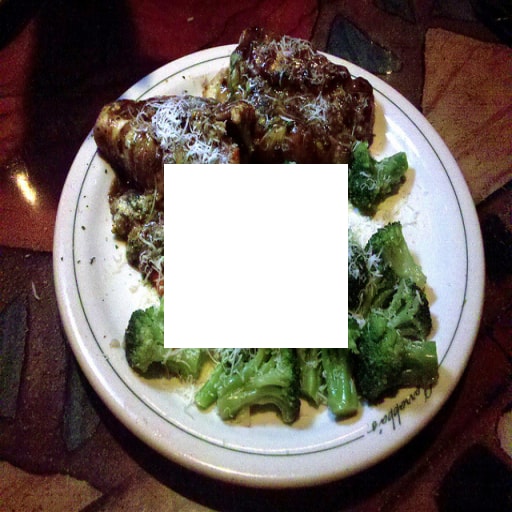}&
  \includegraphics[width=.14\textwidth]{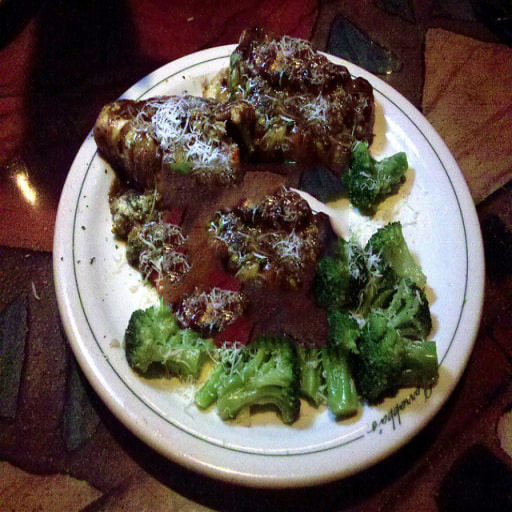}&
  \includegraphics[width=.14\textwidth]{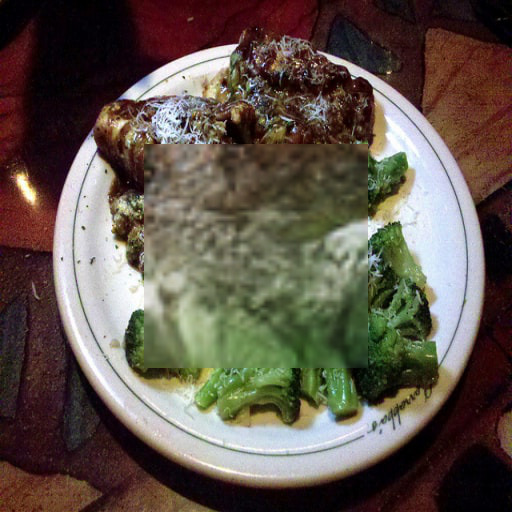}&
  \includegraphics[width=.14\textwidth]{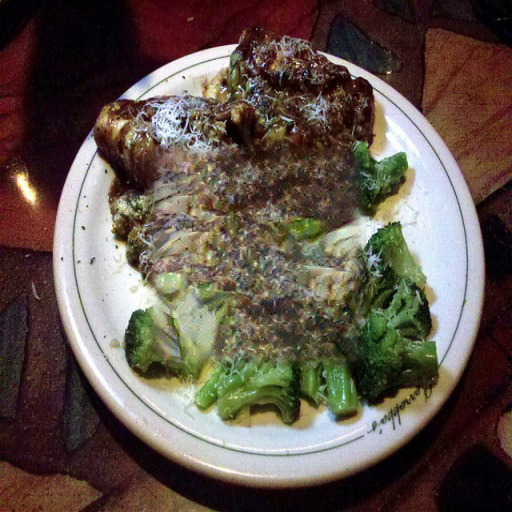}&
  \includegraphics[width=.14\textwidth]{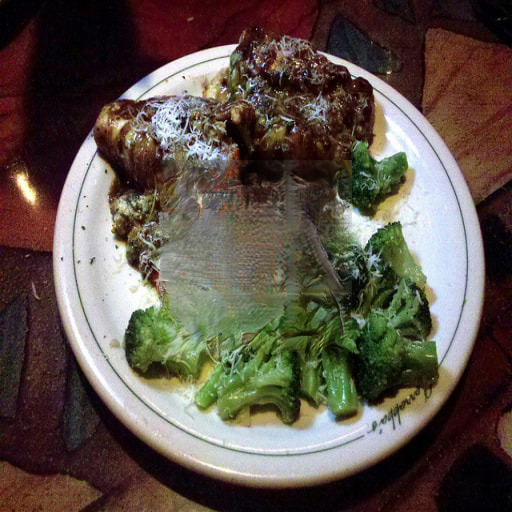}&
  \includegraphics[width=.14\textwidth]{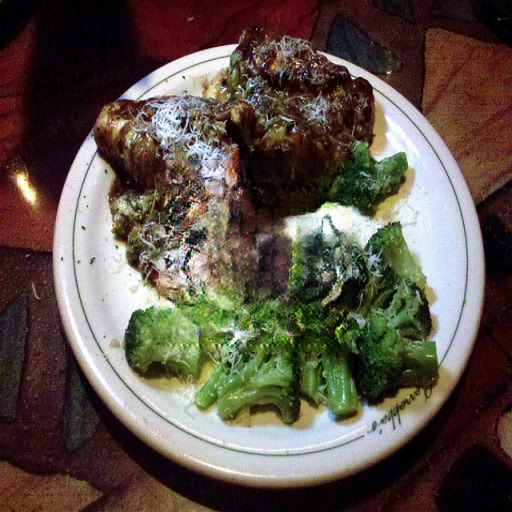}&
  \includegraphics[width=.14\textwidth]{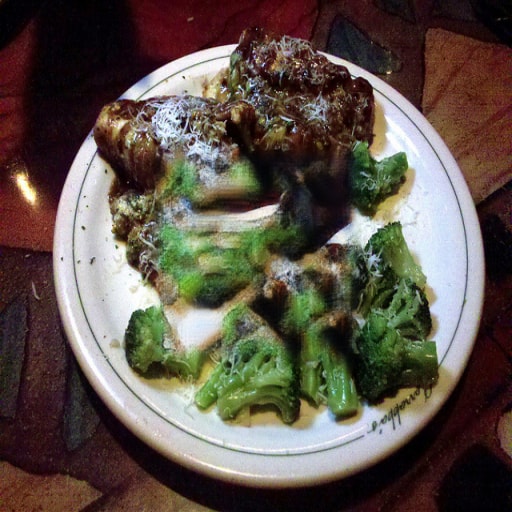}\\
  \includegraphics[width=.14\textwidth]{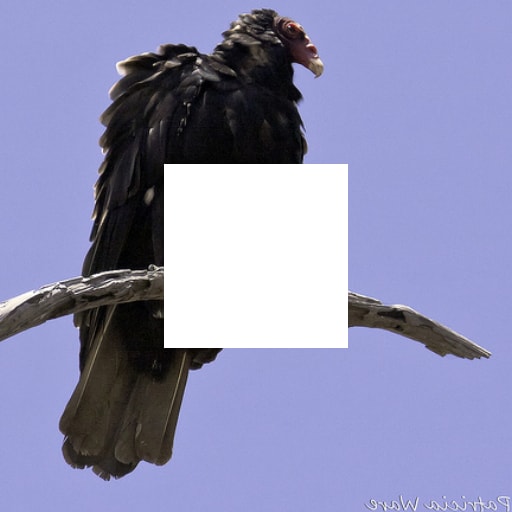}&
  \includegraphics[width=.14\textwidth]{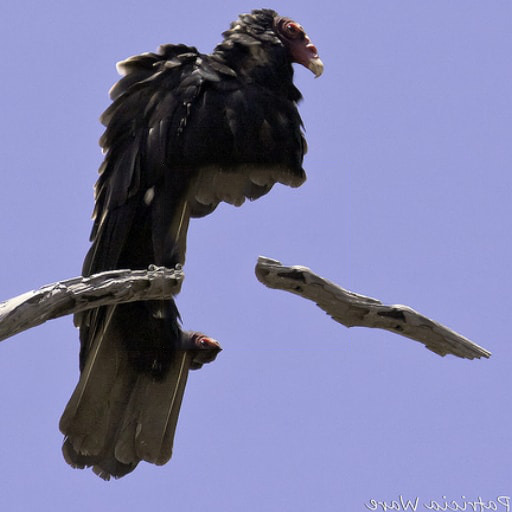}&
  \includegraphics[width=.14\textwidth]{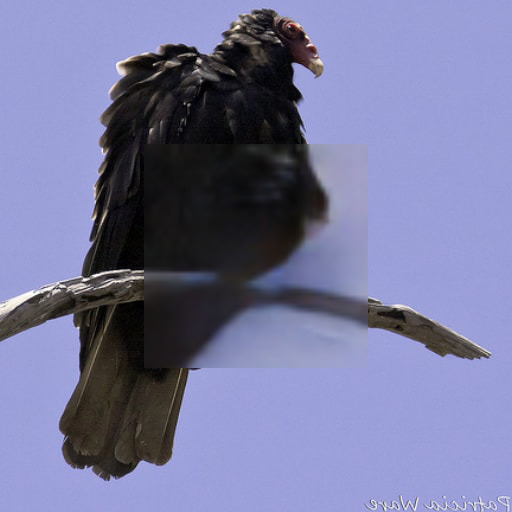}&
  \includegraphics[width=.14\textwidth]{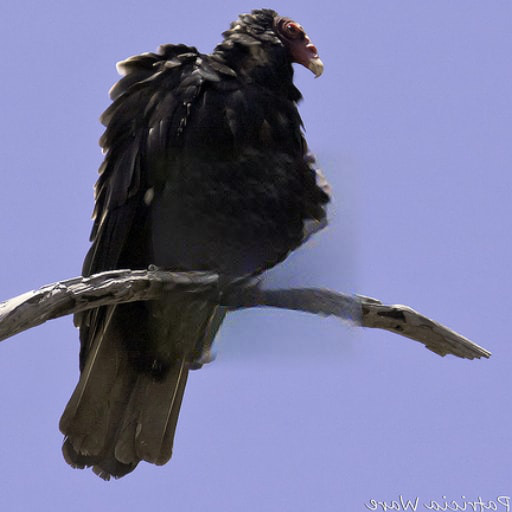}&
  \includegraphics[width=.14\textwidth]{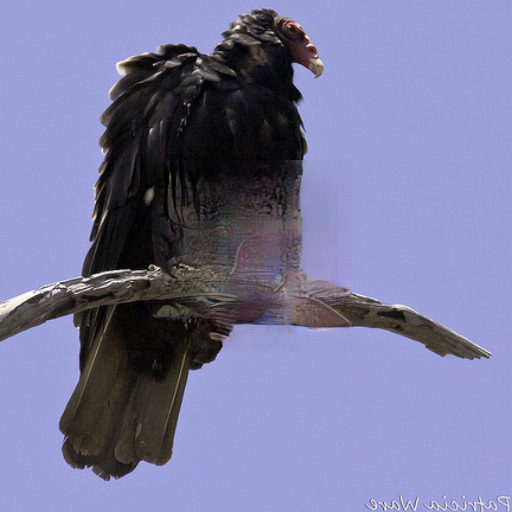}&
  \includegraphics[width=.14\textwidth]{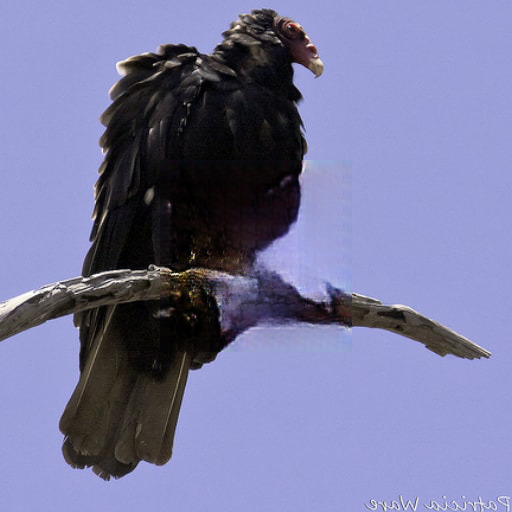}&
  \includegraphics[width=.14\textwidth]{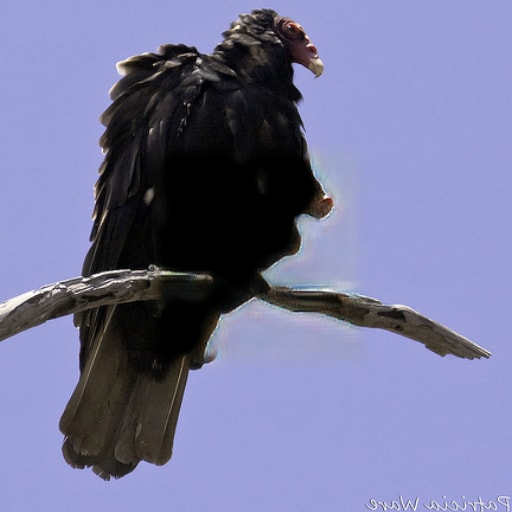}\\
  \includegraphics[width=.14\textwidth]{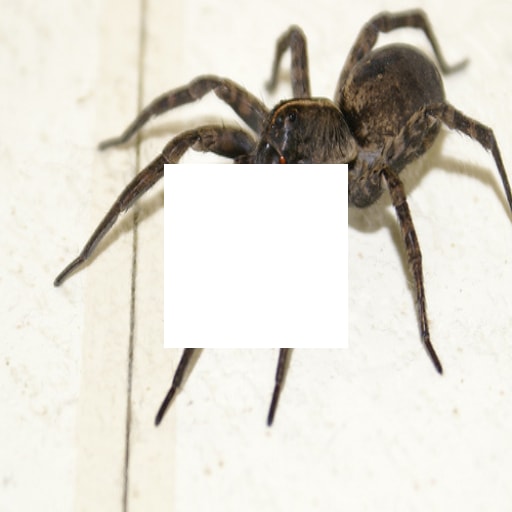}&
  \includegraphics[width=.14\textwidth]{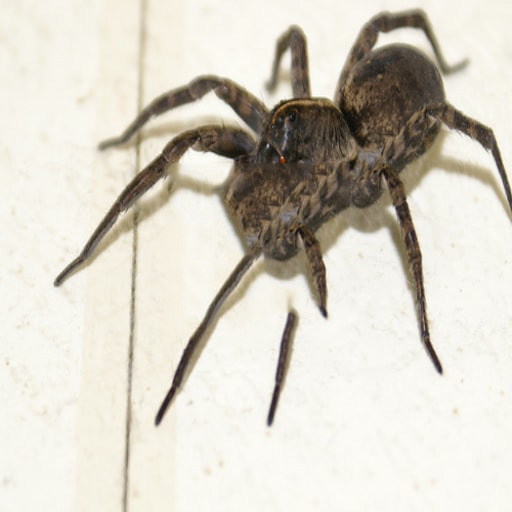}&
  \includegraphics[width=.14\textwidth]{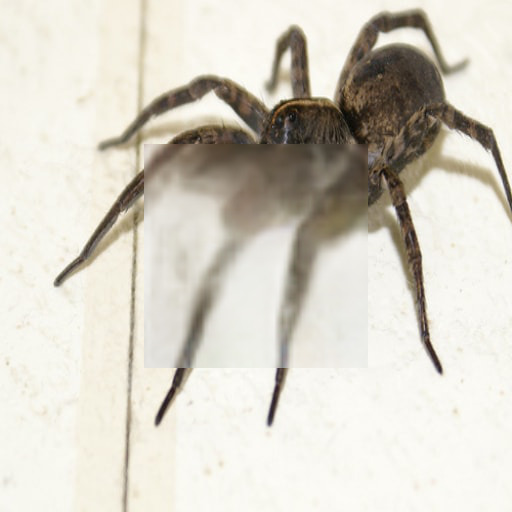}&
  \includegraphics[width=.14\textwidth]{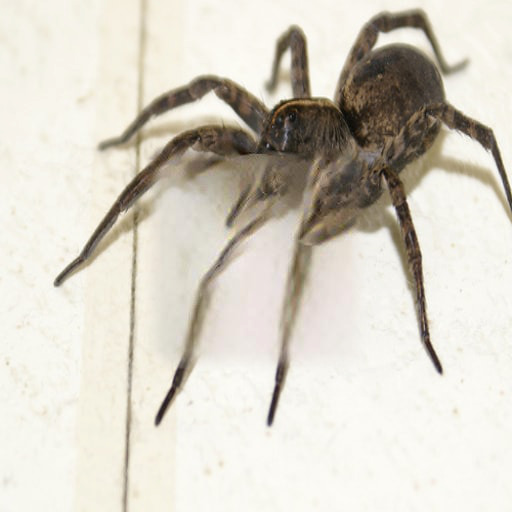}&
  \includegraphics[width=.14\textwidth]{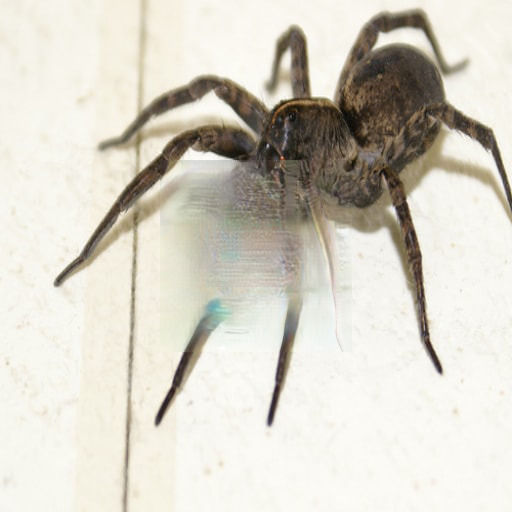}&
  \includegraphics[width=.14\textwidth]{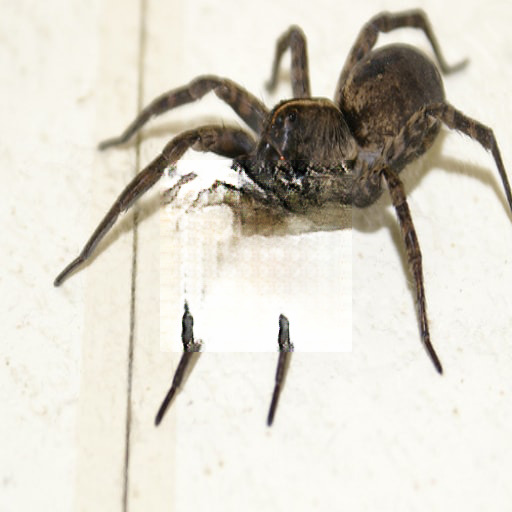}&
  \includegraphics[width=.14\textwidth]{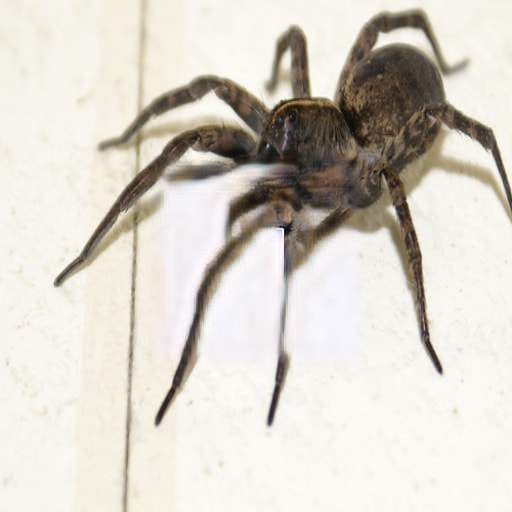}\\
  \includegraphics[width=.14\textwidth]{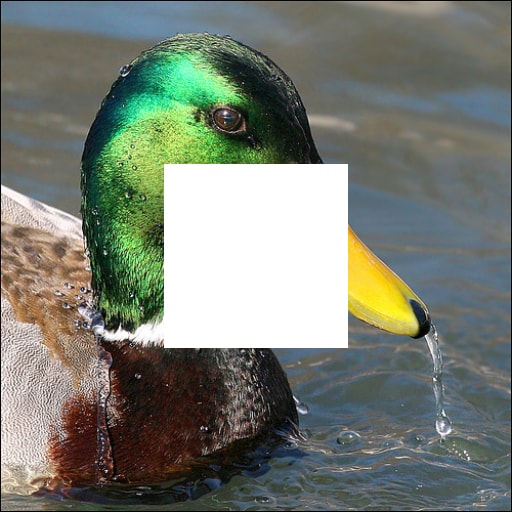}&
  \includegraphics[width=.14\textwidth]{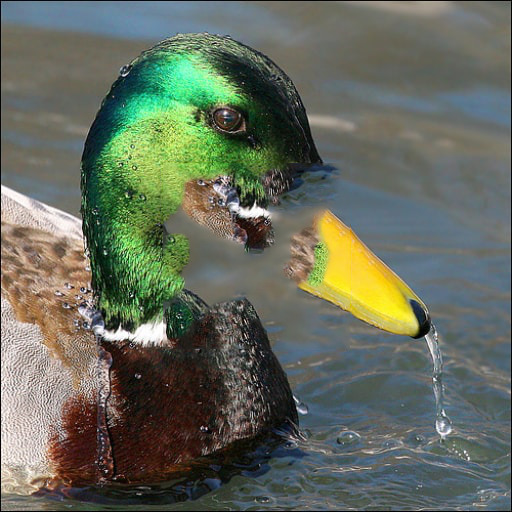}&
  \includegraphics[width=.14\textwidth]{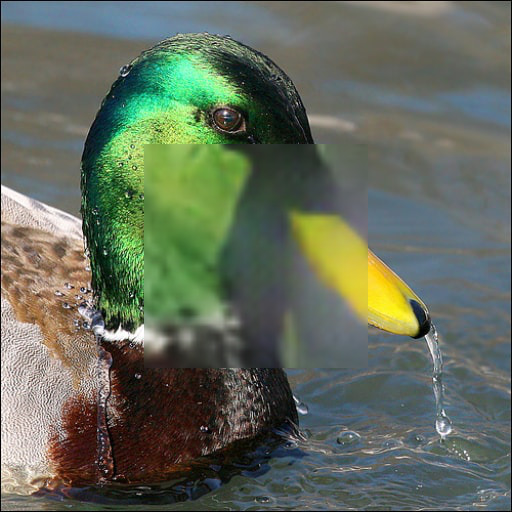}&
  \includegraphics[width=.14\textwidth]{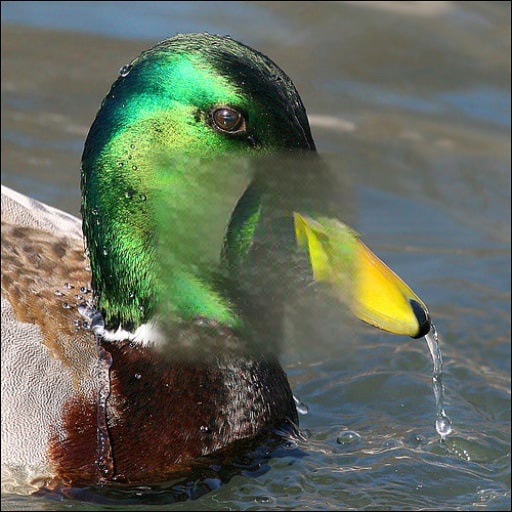}&
  \includegraphics[width=.14\textwidth]{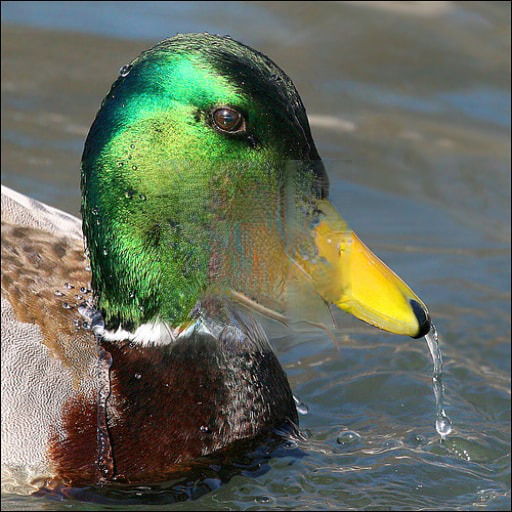}&
  \includegraphics[width=.14\textwidth]{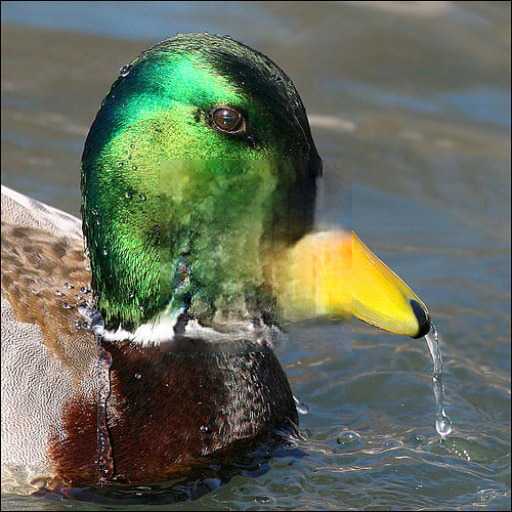}&
  \includegraphics[width=.14\textwidth]{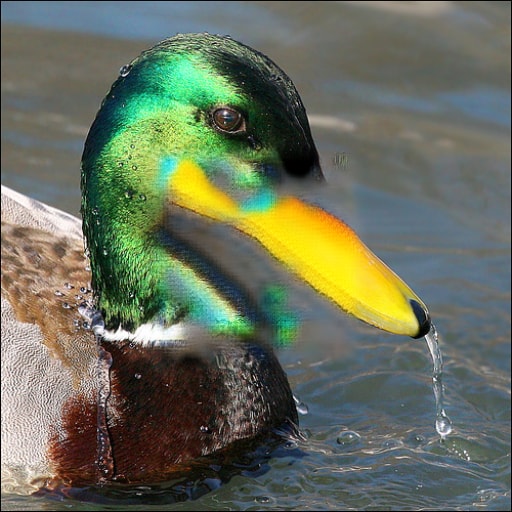}\\
  \includegraphics[width=.14\textwidth]{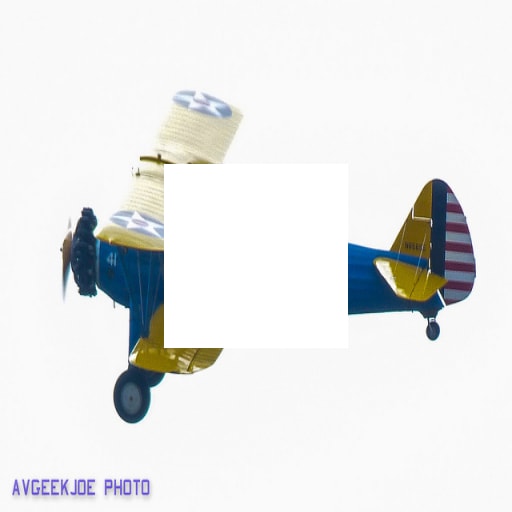}&
  \includegraphics[width=.14\textwidth]{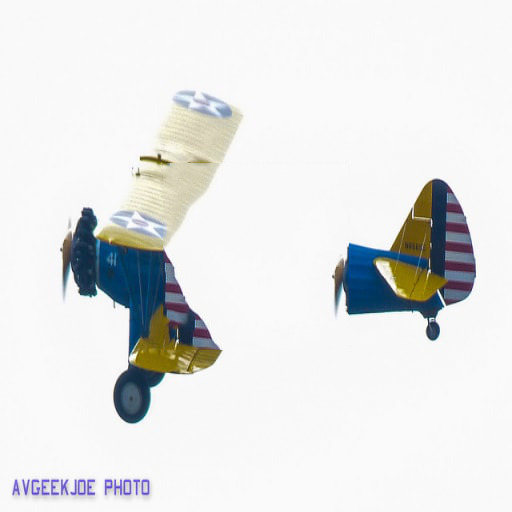}&
  \includegraphics[width=.14\textwidth]{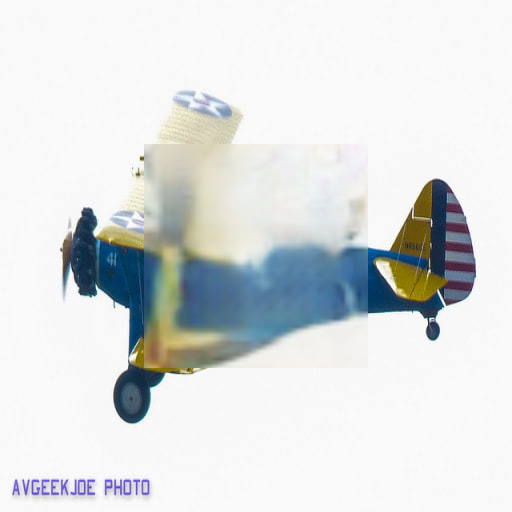}&
  \includegraphics[width=.14\textwidth]{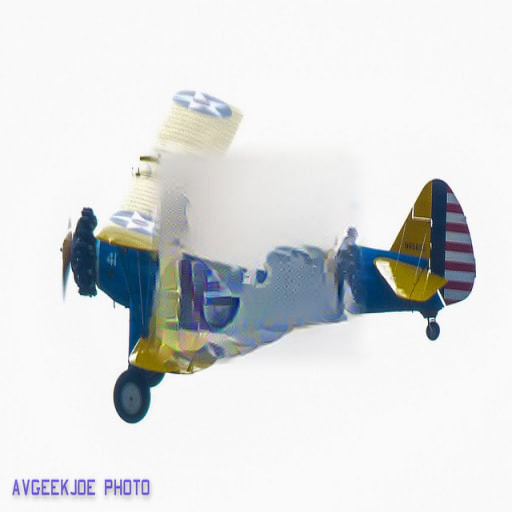}&
  \includegraphics[width=.14\textwidth]{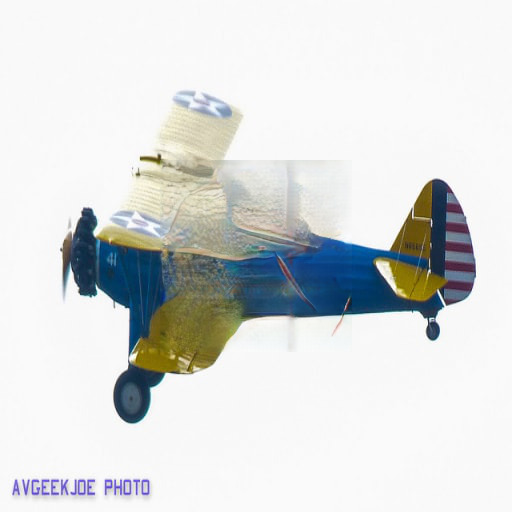}&
  \includegraphics[width=.14\textwidth]{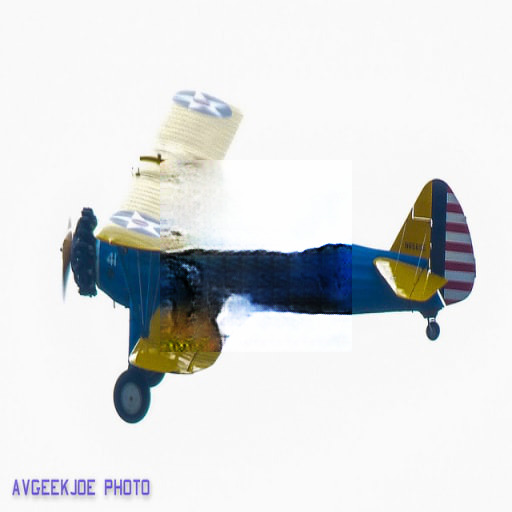}&
  \includegraphics[width=.14\textwidth]{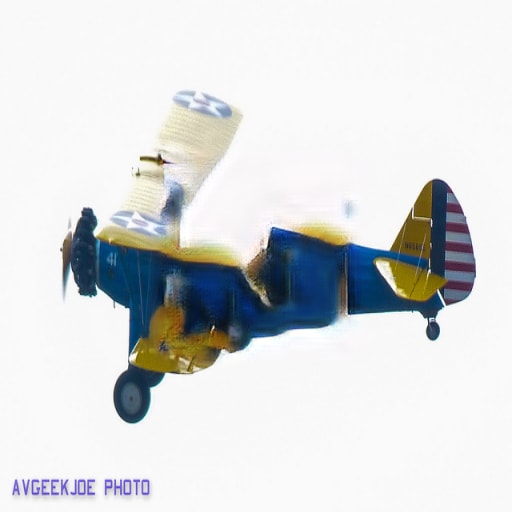}\\
  \includegraphics[width=.14\textwidth]{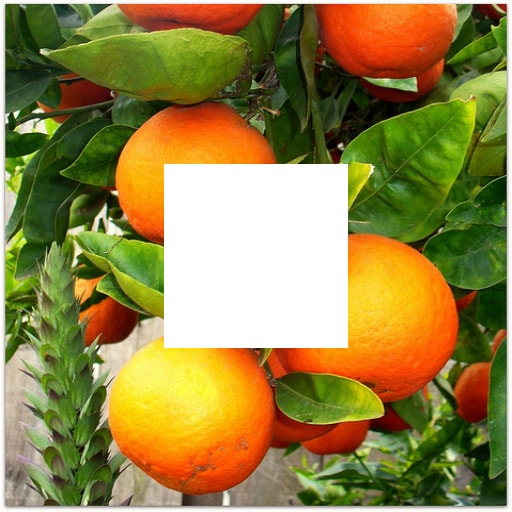}&
  \includegraphics[width=.14\textwidth]{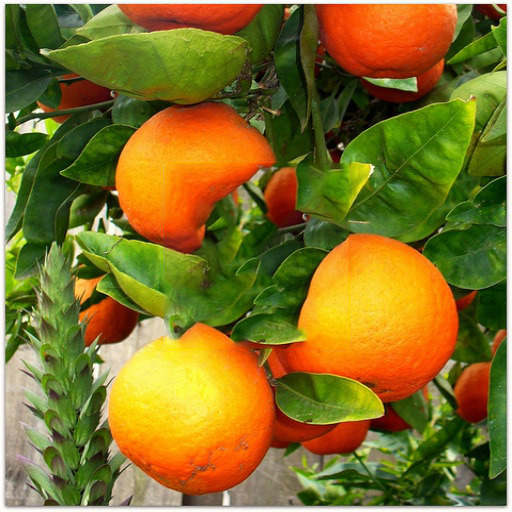}&
  \includegraphics[width=.14\textwidth]{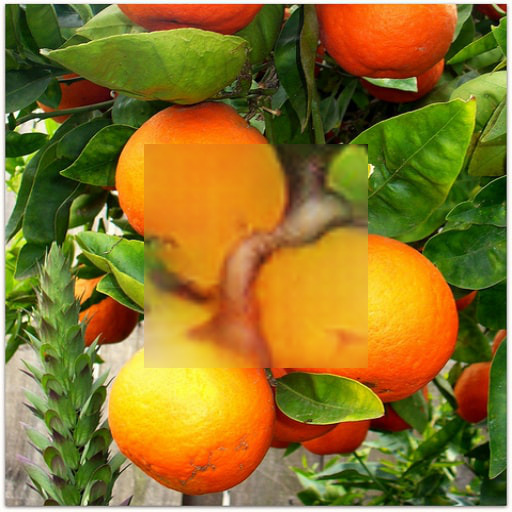}&
  \includegraphics[width=.14\textwidth]{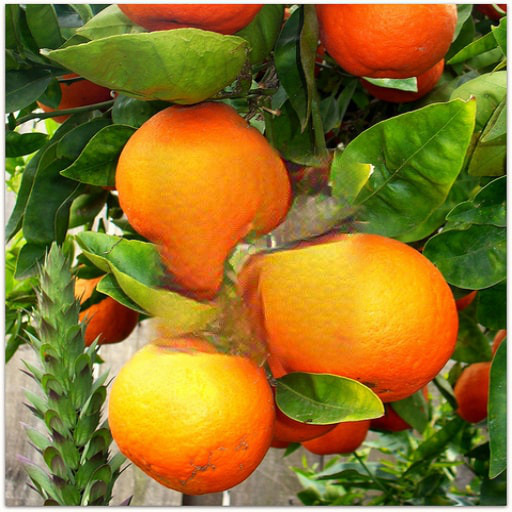}&
  \includegraphics[width=.14\textwidth]{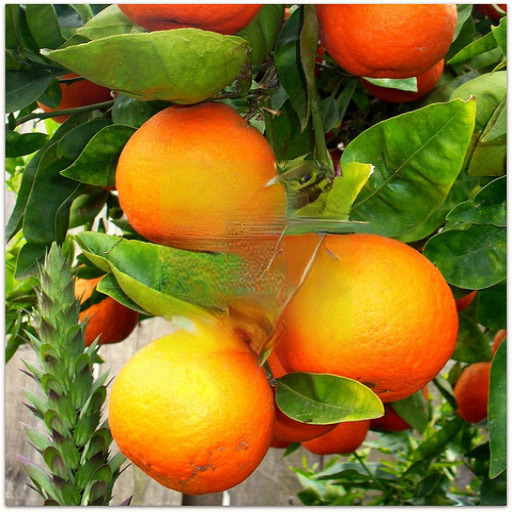}&
  \includegraphics[width=.14\textwidth]{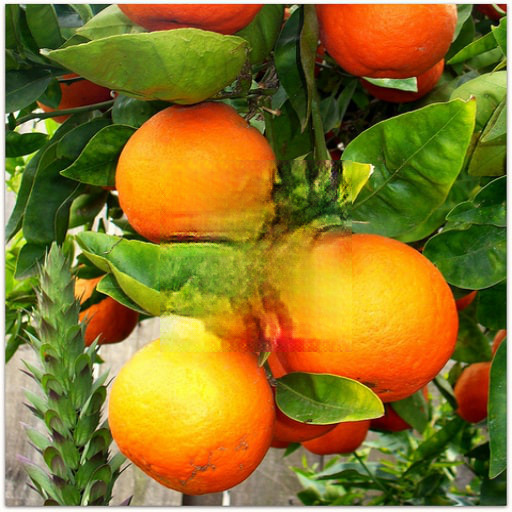}&
  \includegraphics[width=.14\textwidth]{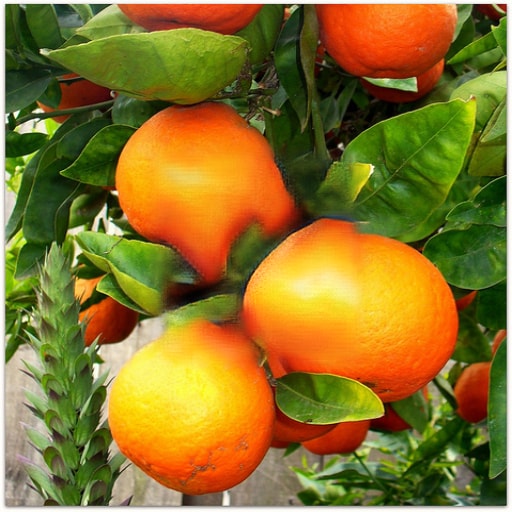}\\

\end{tabular}
\caption{Visual comparisons of ImageNet and COCO results. Each example from left to right: input image, CAF~\cite{barnes2009patchmatch}, CE~\cite{pathak2016context},NPS~\cite{Yang_2017_CVPR}, GLI~\cite{IizukaSIGGRAPH2017}, our result w/o Feature2Image, our final result. All images have size $512\times 512$.}
\label{fig:comparison}
\end{figure*}

\noindent\textbf{Comparison}
Comparing with~\cite{Yang_2017_CVPR}, not only our approach is much faster but also has several advantages. First, the Feature2Image network synthesizes the entire image while~\cite{Yang_2017_CVPR} only optimizes the hole part. By aligning the color of the boundary between the output and the input, we can slightly adjust the tone to make the hole blend with the boundary more seamlessly and naturally (Fig.~\ref{fig:comparison}). Second, our model is trained to directly model the statistics of real-world images and works well on all resolutions, while~\cite{Yang_2017_CVPR} is unable to produce sharp results when the image is small. Comparing with other learning-based inpainting methods, our approach is more general as we can handle larger inputs like 512x512. In contrast, ~\cite{pathak2016context} can only inpaint 128x128 images while ~\cite{IizukaSIGGRAPH2017} is limited to 256x256 images and the holes are limited to be smaller than 128x128. 

\begin{figure}
\centering
\small
\setlength{\tabcolsep}{0.5pt}
\begin{tabular}{ccccccc}
  \includegraphics[width=.14\textwidth]{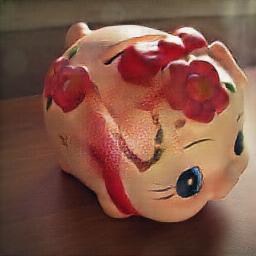}&
  \includegraphics[width=.14\textwidth]{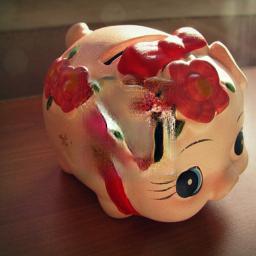}&
  \includegraphics[width=.14\textwidth]{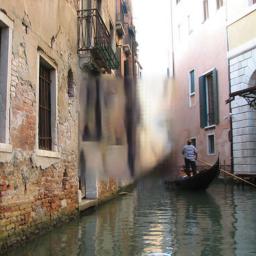}&
  \includegraphics[width=.14\textwidth]{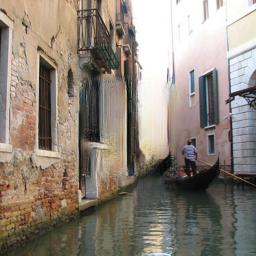}&
  \includegraphics[width=.14\textwidth]{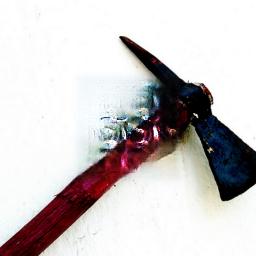}&
  \includegraphics[width=.14\textwidth]{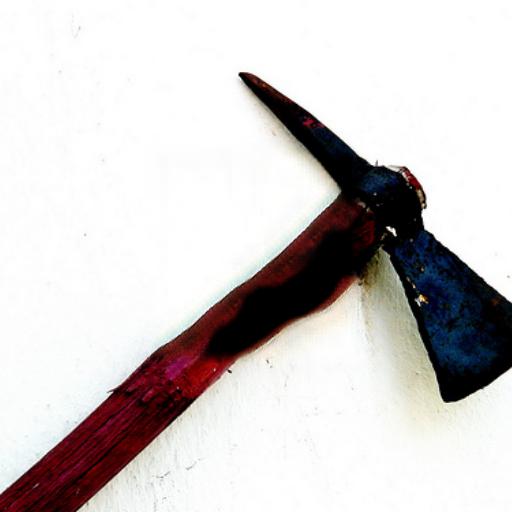}&
  \includegraphics[width=.14\textwidth]{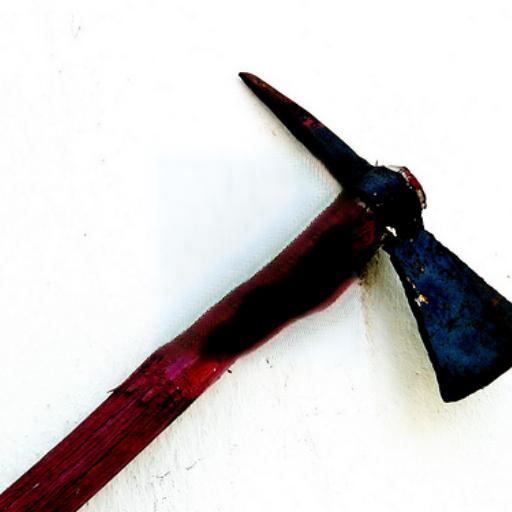}\\
  (a) & (b) & (c) & (d) & (e) & (f) & (g)\\
\end{tabular}
\caption{Left: using deconvolution (a) vs resize-convolution (b). Middle: using $\ell_2$ reconstruction loss (c) vs using perceptual loss (d). Right: Training Feature2Image network using different input data. (e) Result when trained with the Image2Feature prediction. (f) Result when trained with ground truth. (g) Result when fine-tuned with ground truth and prediction mixtures.}
\label{fig:ablation}
\end{figure}

\noindent\textbf{Ablation Study}
For the Feature2Image network, we observed that replacing the deconvolutional layers in the decoder part with resize-convolution layers resolves the checkerboard patterns as described in~\cite{odena2016deconvolution} (Fig.~\ref{fig:ablation} left). We also tried only using $\ell_2$ loss instead of perceptual loss, which gives blurrier inpainting (Fig.~\ref{fig:ablation} middle). Additionally, we experimented different activation layers of VGG19 to extract features and found that \emph{relu3\_1} works better than \emph{relu2\_1} and \emph{relu4\_1}.  

We may also use iterative inference by running Feature2Image network multiple times. At each iteration, the final output is used as input to VGG and patch-swap, and then again given to Feature2Image network for inference. We found iteratively applying Feature2Image improves the sharpness of the texture but sometimes aggregates the artifacts near the boundary.

For the Image2Feature network, an alternative is to use vanilla context encoder~\cite{pathak2016context} to generate $I^0_0$ as the initial inference. However, we found our model produces better results as it is much deeper, and leverages the fully convolutional network and dilated layer.

As discussed in Sec.~\ref{sec:training}, an important practice to guarantee successful training of the Feature2Image network is to use ground truth image as input rather than using the output of the Image2Feature network. Fig.~\ref{fig:ablation} also shows that training with the prediction from the Image2Feature network gives very noisy results, while the models trained with ground truth or further fine-tuned with ground-truth and prediction mixtures can produce satisfying inpainting.

\begin{figure}
\centering
\small
\setlength{\tabcolsep}{1pt}
\begin{tabular}{cccccc}
  \includegraphics[width=.16\textwidth]{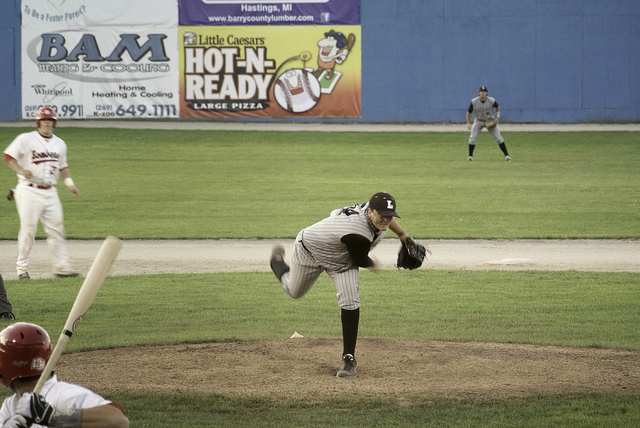}&
  \includegraphics[width=.16\textwidth]{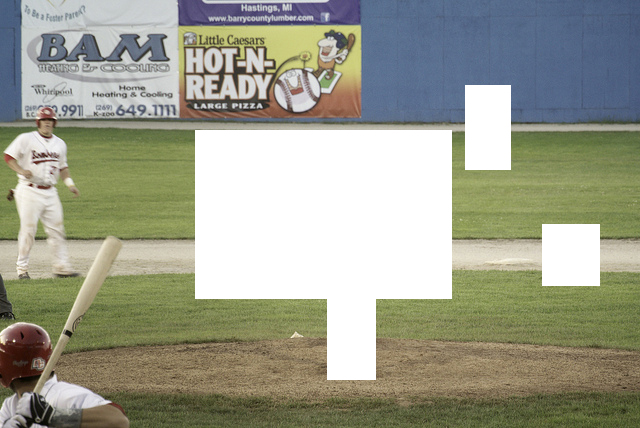}&
  \includegraphics[width=.16\textwidth]{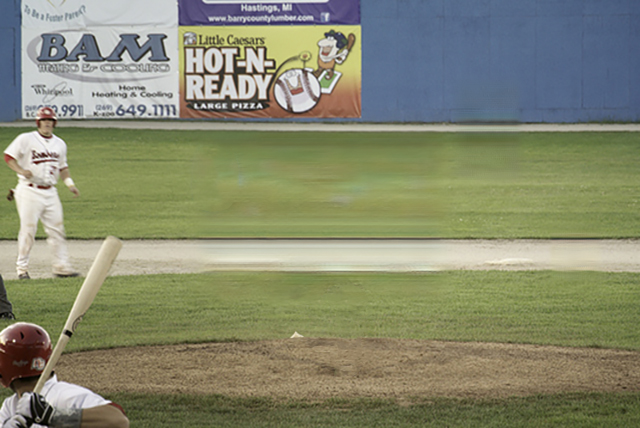}&
  \includegraphics[width=.16\textwidth]{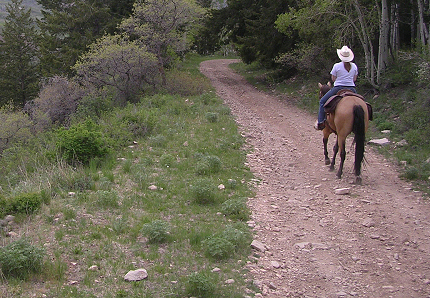}&
  \includegraphics[width=.16\textwidth]{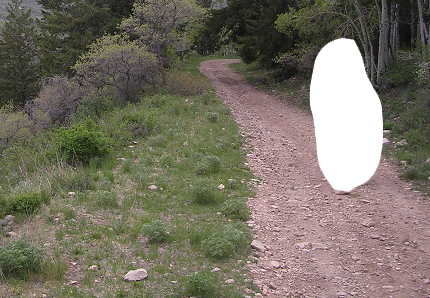}&
  \includegraphics[width=.16\textwidth]{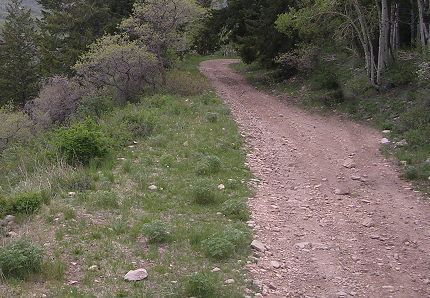}\\
  (a) & (b) & (c) & (d) & (e) & (f)
\end{tabular}
\caption{Arbitrary shape inpainting of real-world photography. (a), (d): Input. (b), (e): Inpainting mask. (c), (f): Output.}
\label{fig:arbitrary}
\end{figure}

Our framework can be easily applied to real-world tasks. Fig.~\ref{fig:arbitrary} shows examples of using our approach to remove unwanted objects in photography. Given our network is fully convolutional, it is straight-straightforward to apply it to photos of arbitrary sizes. It is also able to fill in holes of arbitrary shapes, and can handle much larger holes than~\cite{iizuzuka2017globally}. 

\begin{figure}
\centering
\small
\setlength{\tabcolsep}{1pt}
\begin{tabular}{cccccc}
  \includegraphics[width=.16\textwidth]{figures/singlescale/boat_256.jpg}&
  \includegraphics[width=.16\textwidth]{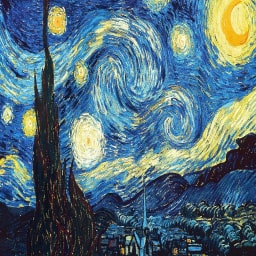}&
  \includegraphics[width=.16\textwidth]{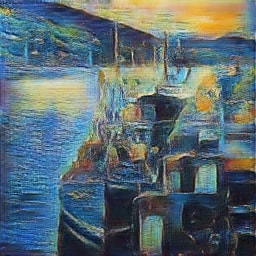}&
  \includegraphics[width=.16\textwidth]{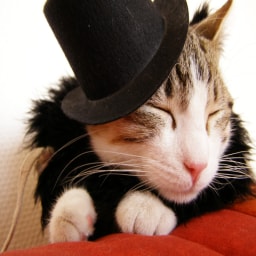}&
  \includegraphics[width=.16\textwidth]{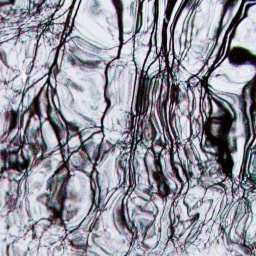}&
  \includegraphics[width=.16\textwidth]{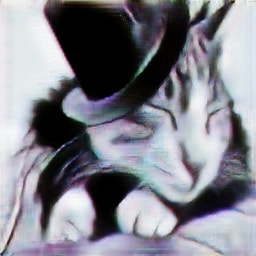}\\
  (a) & (b) & (c) & (d) & (e) & (f)\\
\end{tabular}
\caption{Arbitrary style transfer. (a), (d): Content. (b), (e): Style. (c), (f): Result.}
\label{fig:othertasks}
\end{figure}

The Feature2Image network essentially learns a universal function to reconstruct an image from a swapped feature map, therefore can also be applied to other tasks. For example, by first constructing a swapped feature map from a content and a style image, we can use the network to reconstruct a new image for style transfer. Fig.~\ref{fig:othertasks} shows examples of using our Feature2Image network trained on COCO towards arbitrary style transfer. Although the network is agnostic to the styles being transferred, it is still capable of generating satisfying results and runs in real-time. This shows the strong generalization ability of our learned model, as it's only trained on a single COCO dataset, unlike other style transfer methods.

\begin{figure}
\centering
\small
\setlength{\tabcolsep}{1pt}
\begin{tabular}{cccccc}
  \includegraphics[width=.16\textwidth]{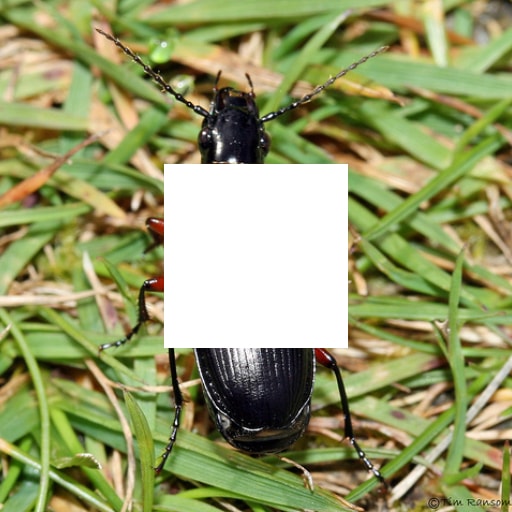}&
  \includegraphics[width=.16\textwidth]{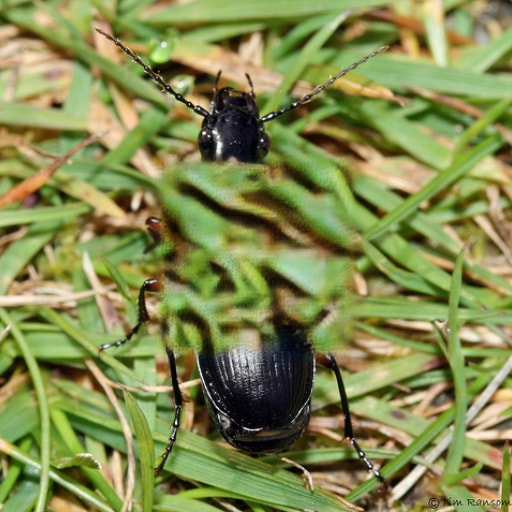}&
  \includegraphics[width=.16\textwidth]{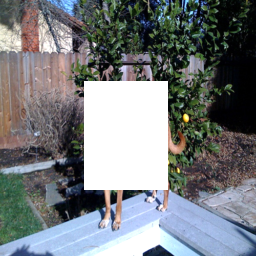}&
  \includegraphics[width=.16\textwidth]{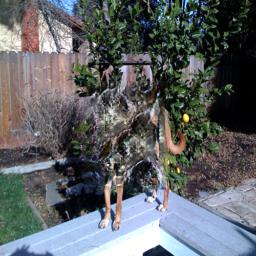}&
  \includegraphics[width=.16\textwidth]{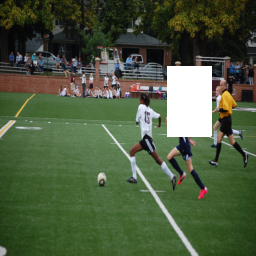}&
  \includegraphics[width=.16\textwidth]{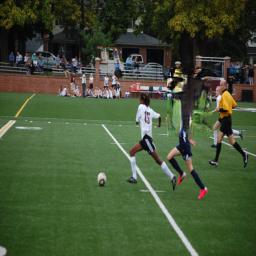}\\
  (a) & (b) & (c) & (d) & (e) & (f)\\
\end{tabular}
\caption{Failure cases. (a), (c) and (e): Input. (b), (d) and (f): Output.}
\label{fig:failure}
\end{figure}

Our approach is very good at recovering a partially missing object like a plane or a bird (Fig.~\ref{fig:comparison}). However, it can fail if the image has overly complicated structures and patterns, or a major part of an object is missing such that Image2Feature network is unable to provide a good inference (Fig.~\ref{fig:failure}).

\section{Conclusion}

We propose a learning-based approach to synthesize missing contents in a high-resolution image. Our model is able to inpaint an image with realistic and sharp contents in a feed-forward manner. We show that we can simplify training by breaking down the task into multiple stages, where the mapping function in each stage has smaller dimensionality. It is worth noting that our approach is a meta-algorithm and naturally we could explore a variety of network architectures and training techniques to improve the inference and the final result. We also expect that similar idea of multi-stage, multi-scale training could be used to directly synthesize high-resolution images from sampling.

\section{Acknowledgments}
This work was supported in part by the ONR YIP grant N00014-17-S-FO14, the CONIX Research Center, one of six centers in JUMP, a Semiconductor Research Corporation (SRC) program sponsored by DARPA, the Andrew and Erna Viterbi Early Career Chair, the U.S. Army Research Laboratory (ARL) under contract number W911NF-14-D-0005, Adobe. The content of the information does not necessarily reflect the position or the policy of the Government, and no official endorsement should be inferred.

\clearpage

\bibliographystyle{splncs}
\bibliography{egbib}
\end{document}